\documentclass[10pt,journal]{IEEEtran}
\usepackage{graphicx}
\usepackage{amsmath,amssymb}
\usepackage{multirow}
\usepackage{algorithm}
\usepackage{algorithmicx}
\usepackage{algpseudocode}
\usepackage{makecell}
\usepackage{blkarray}
\usepackage{url}
\usepackage{subfig}
\usepackage{diagbox}  
\usepackage{xcolor}
\usepackage{colortbl,booktabs}
\usepackage{color}  
\newcommand{\red}[1]{\textcolor{red}{#1}}  
\newcommand{\blue}[1]{\textcolor{black}{#1}}
\newcommand{\TheName}{\emph{SenseMag}}
\title{\TheName: Enabling Low-Cost Traffic Monitoring using Non-invasive Magnetic Sensing}

\begin{document}
%

\author{Kafeng Wang, Haoyi Xiong, \emph{Member, IEEE,} Jie Zhang, \emph{Member, IEEE,}  Hongyang Chen, \emph{Senior Member, IEEE,} Dejing Dou, \emph{Senior Member, IEEE,} and Cheng-Zhong Xu,~\emph{Fellow, IEEE}~\thanks{
K. Wang is with Shenzhen Institute of Advanced Technology, Chinese Academy of Sciences, and University of Chinese Academy of Sciences, Shenzhen, Guangdong, China (email: kf.wang@siat.ac.cn). 
H. Xiong and Dejing Dou are with Big Data Lab, Baidu Inc., Haidian, Beijing, China. J. Zhang is with Key Laboratory of High Confidence Software Technologies, Peking University, Haidian, Beijing, China (email: xionghaoyi@baidu.com). 
H. Chen is with Research Center for Intelligent Network, Zhejiang Lab, Hangzhou, Zhejiang, China  (email: dr.h.chen@ieee.org). 
C.Z. Xu is with  State Key Laboratory of Internet of Things for Smart City, and Department of Computer and Information Science, University of Macau, Tapia, Macau, China (email: czxu@um.edu.mo). 
Kafeng Wang is also with Guangdong-Hong Kong-Macao Joint Laboratory of Human-Machine Intelligence-Synergy Systems, Shenzhen Institute of Advanced Technology, Chinese Academy of Sciences, Shenzhen, 518055, China.
Please contact to Dr. Haoyi Xiong and Prof. Cheng-Zhong Xu for correspondence.
} }

\maketitle




\begin{abstract}
The operation and management of intelligent transportation systems (ITS), such as traffic monitoring, relies on real-time data aggregation of vehicular traffic information, including vehicular types (e.g., cars, trucks, and buses), in the critical roads and highways. While traditional approaches based on vehicular-embedded GPS sensors or camera networks would either invade drivers' privacy or require high deployment cost, this paper introduces a low-cost method, namely \TheName, to recognize the vehicular type using a pair of non-invasive magnetic sensors deployed on the straight road section. \TheName\ filters out noises and segments received magnetic signals by the exact time points that the vehicle arrives or departs from every sensor node. Further, \TheName\ adopts a hierarchical recognition model to first estimate the speed/velocity, then identify the length of vehicle using the predicted speed, sampling cycles, and the distance between the sensor nodes. With the vehicle length identified and the temporal/spectral features extracted from the magnetic signals, \TheName\ classify the types of vehicles accordingly. Some semi-automated learning techniques have been adopted for the design of filters, features, and the choice of hyper-parameters. Extensive experiment based on real-word field deployment (on the highways in Shenzhen, China) shows that \TheName\ significantly outperforms the existing methods in both classification accuracy and the granularity of vehicle types (i.e., 7 types by \TheName\ versus 4 types by the existing work in comparisons). To be specific, our field experiment results validate that \TheName\ is with at least $90\%$ vehicle type classification accuracy and less than 5\% vehicle length classification error.

\end{abstract}



\section{Introduction}
With the rapid development of ubiquitous sensing, communication and computing devices, smart cities with Intelligent Transportation Systems (ITS)~\cite{zhang2011data} are conforming to new standards and requirements of modern urbanization and civilization. The smart operation and management of ITS, such as the monitoring of urban traffic safety, relies on real-time data aggregation of vehicular traffic information in the critical roads and highways of the city. For example, to smooth the urban traffic in rush hour, ITS frequently encourages passengers to share drives in certain roads/slots through car pooling. To specify the roads for car pooling, ITS needs to measure the traffic volume, vehicular type (private cars or public buses), and traffic status of each key road in the rush hours. While the traffic status (e.g., congested/jammed/smooth) can be identified by the speed of vehicles, there thus needs a method to identify and do statistics on the vehicular types and the traffic speed on the roads~\cite{tian2014hierarchical}.

\begin{figure}
\centering
\includegraphics[width=0.5\textwidth]{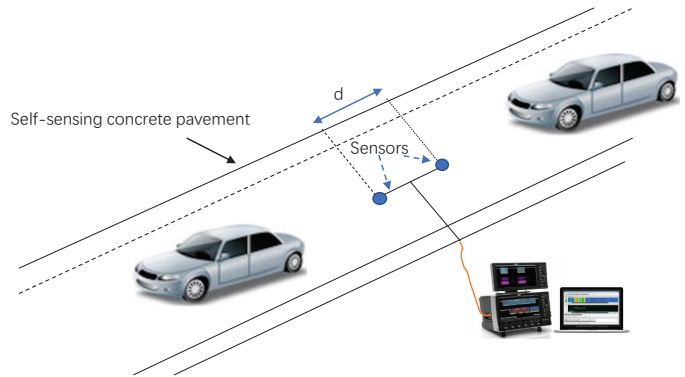}
\caption{Deployment of Magnetic Sensors for Traffic Monitoring. (Specifically, \TheName\ adopts two magnetic sensors to gather information.)}
\label{fig:deployment}
\end{figure}

Generally, we can categorize the existing methods to obtain aforementioned information in two folders: infrastructure-based approach~\cite{miller2008vehicle,milanes2012intelligent} and infrastructure-less approach~\cite{el2018towards,silva2013traffic}. To build an infrastructure for traffic monitoring, traditional method is to use camera array networks, where a large number of cameras are deployed to sense every corner of the streets/roads/highways to visually track each moving vehicle. They can localize each vehicle, identify its type, and measure its speed all using computer vision techniques. In terms of infrastructure-less approach, one can obtain the information about vehicular type and speed using vehicular-embedded GPS sensors in a so-called crowdsensing manner~\cite{zhang20144w1h,xiong2016sensus,xiong2015crowdtasker}. For example, some navigation Apps~\cite{silva2013traffic,hofstede2009surfmap} on smartphones would first require drivers to input their personal information, then track their real-time mobility, including GPS location and speed. With the large-scale GPS tracking and data aggregation, one can easily map and update the statistics on the vehicular types and speed onto each road of the city in real-time. Note that GPS-based solution no longer works when the user turns off the mobility tracking, while the deployment of camera networks relies on the huge monetary investment. Further, both existing infrastructure-based and infrastructure-less techniques would invade the privacy of drivers. All above issues would burden the scalability of traffic monitoring. Thus a low cost method is needed to obtain vehicular type and speed in real-time without seriously invade users' privacy~\cite{burney2017google}.

\begin{figure*}
\centering
\subfloat[Original Temporal]{\includegraphics[width=0.2\textwidth, height=0.4\textheight]{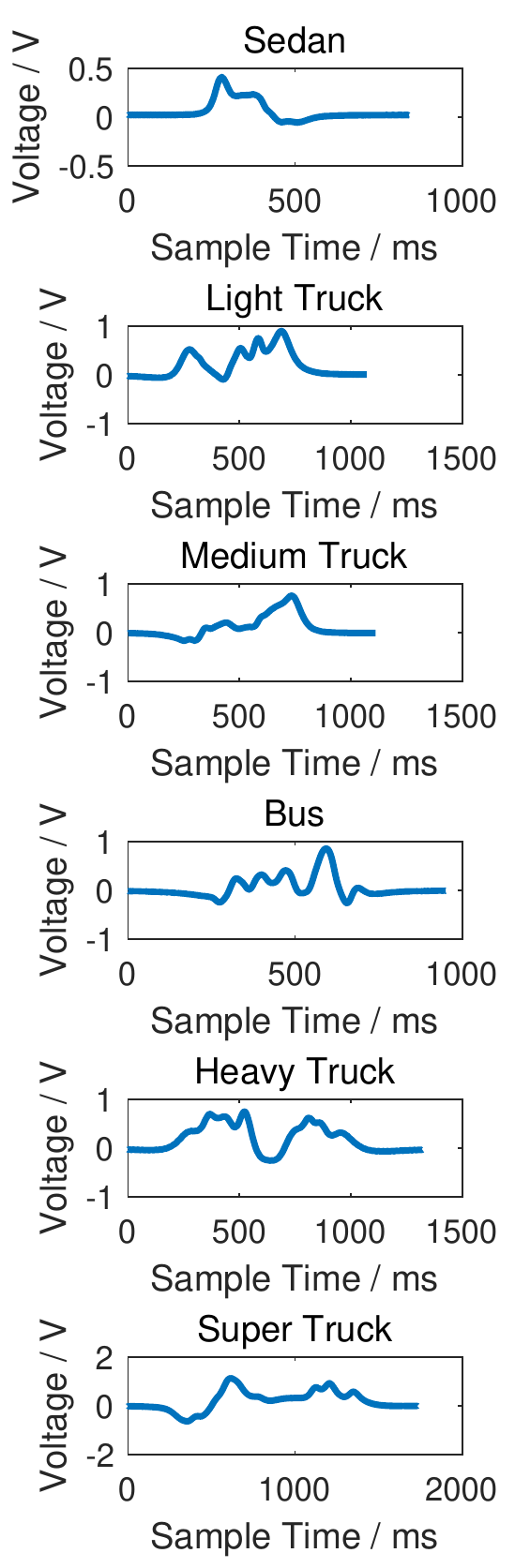}}\
\subfloat[Original Frequency]{\includegraphics[width=0.2\textwidth, height=0.4\textheight]{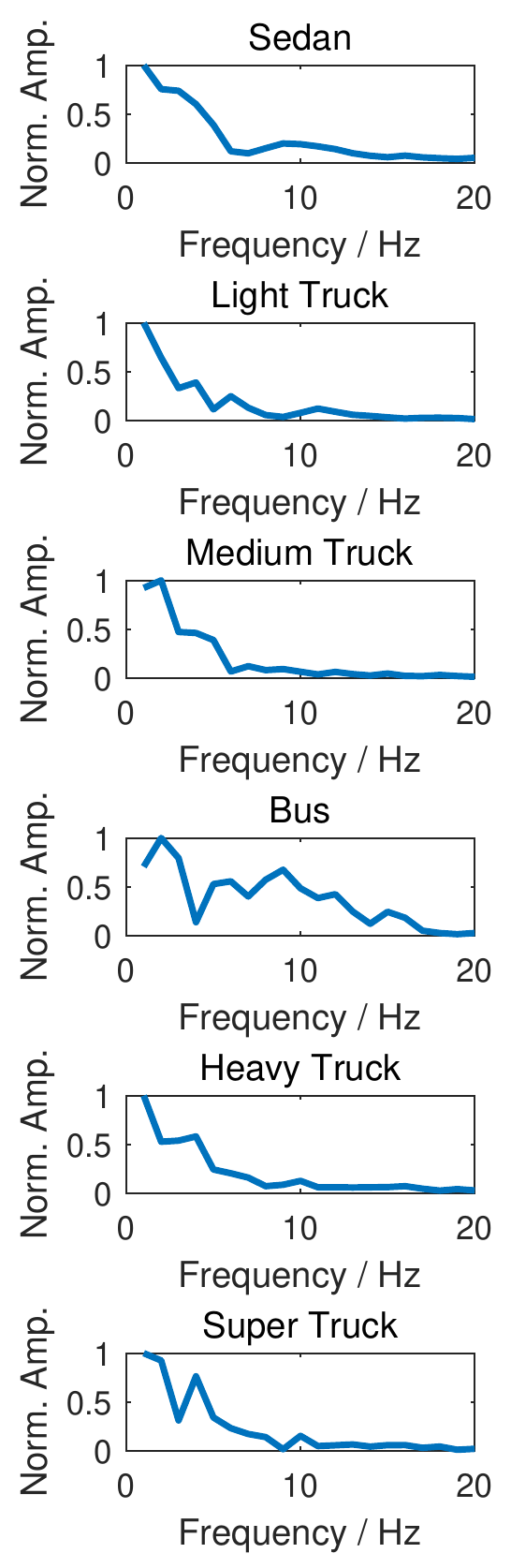}}\
\subfloat[Processed Temporal]{\includegraphics[width=0.2\textwidth, height=0.4\textheight]{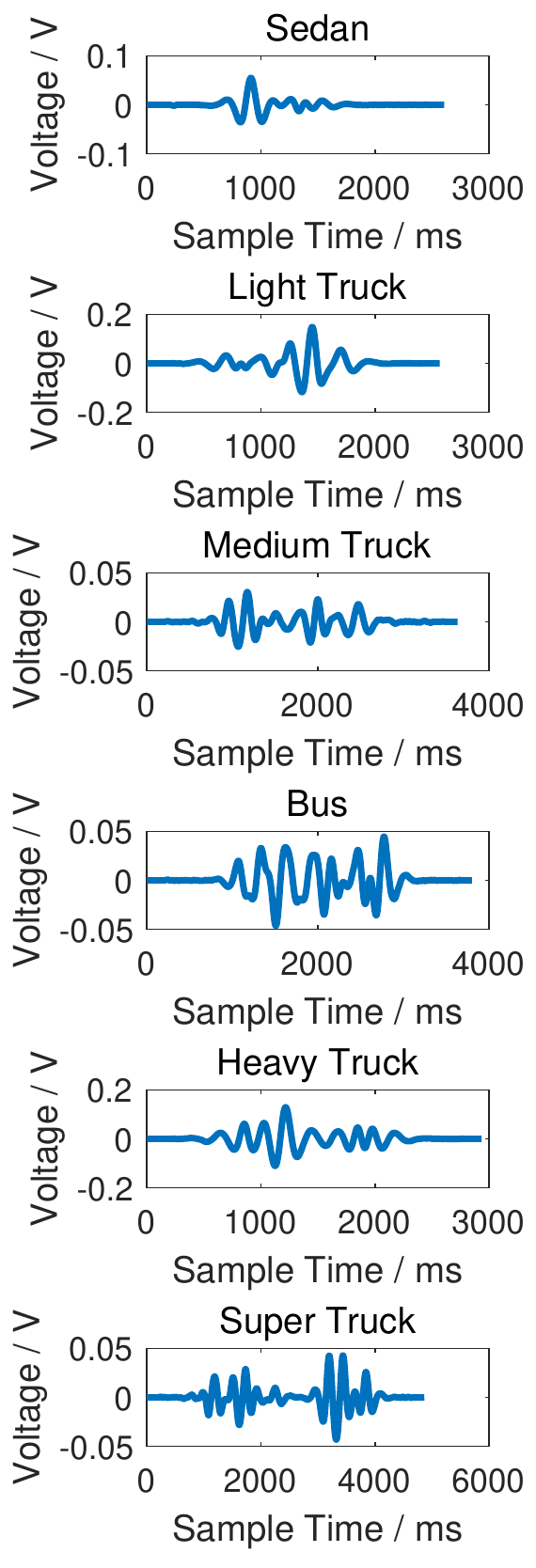}}\
\subfloat[Processed Frequency]{\includegraphics[width=0.2\textwidth, height=0.4\textheight]{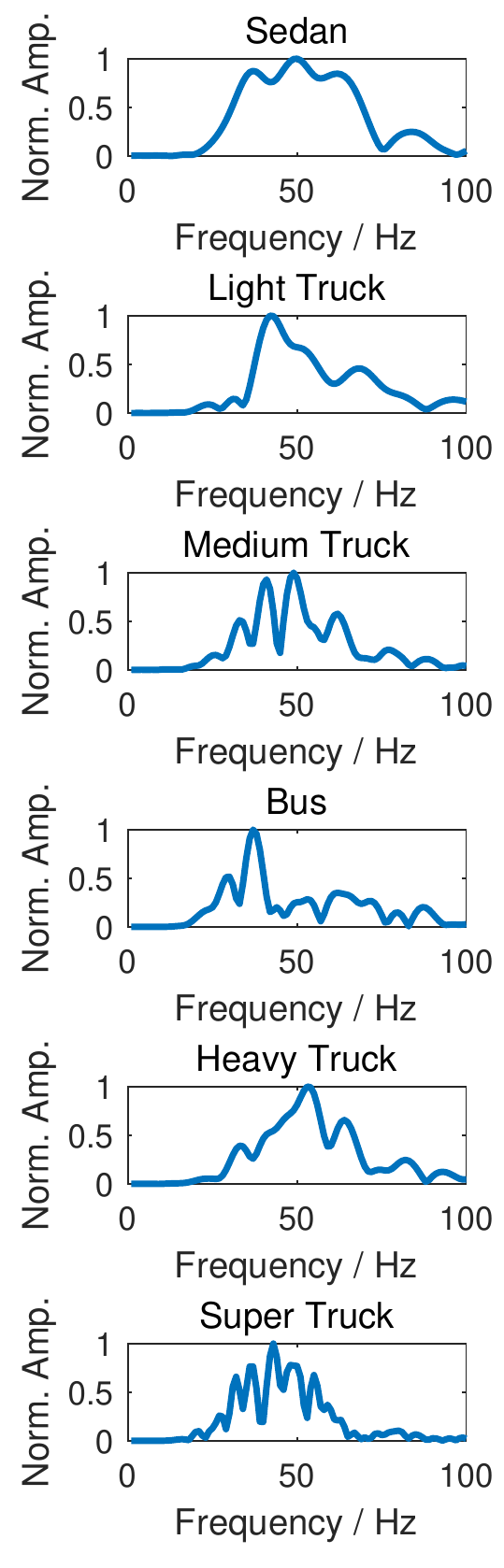}}\
\caption{Magnetic Signals in Temporal and Frequency Domains (The detailed way that we process the signal representations in the frequency domain will be introduced in Section IV.B). 
}
\label{fig:proc_temp_freq}
\end{figure*}

Among a wide range of approaches, magnetic sensors can identify the vehicular types and measure the speed without tracking each individual users' mobility in an non-invasive manner~\cite{Caruso1999Vehicle,Cheung2007Traffic,zhang2010distributed,zhang2011real,Xu2016Traffic,xu2017vehicle,xu2018vehicle,feng2020magmonitor}. Figure~\ref{fig:deployment} demonstrates a typical deployment of magnetic sensors for traffic monitoring. Prior to the traffic monitoring, a magnetic sensing system that can detect the change of surrounding magnetic fields has been already deployed on/under the surface of highway. Specifically, any approaching vehicle can be viewed as a moving magnet that would cause slight perturbation in the surrounding magnetic field. In this way, per vehicle approaching, the magnetic sensing system would be activated by the change of magnetic field and start to collect magnetic field signals for further processing and recognition. For example, Figure~\ref{fig:proc_temp_freq}(a) presents the magnetic signals with various approaching vehicles, where the magnetic signals are represented as the changes/trends of output voltages of the magnetic field sensors over time (in milliseconds). We could observe the patterns of signals and their difference for various vehicle types. With the collection of labeled data, such as clips of magnetic signals recorded with identified vehicular types and speed, one can use general purpose supervised learning techniques to identify the vehicles and classify the vehicular type. Such method works but is with poor performance.

\begin{table}
\caption{Vehicle Types and the Standard Ranges of Vehicle Lengths in China}
\label{table:vehicle_classification}
\centering
\begin{tabular}{|l|c|c|}
\hline
Vehicle Type & Length Interval (m)& Axles \\
\hline
Motorbike & $(0,3]$ & 0 \\
\hline
Sedan, SUV & $(3,6]$ & 2 \\
\hline
Light truck & $(3,6]$ & 2 \\
\hline
Medium truck & $(6,12]$ & 2 \\
\hline
Bus & $(6,12]$ & 2 \\
\hline
Heavy truck & $(6,12]$ & 3 or 4 \\
\hline
Super truck & $(12,20]$ & 4 or more\\
\hline
\end{tabular}
\end{table}

To categorize vehicles by their types, such as Sedan and SUV, Bus, and Light/Medium/Heavy/Super Truck, there frequently needs information on two attributes---the length of vehicles and the number of axles~\cite{minge2012loop,TPRI}. Table~\ref{table:vehicle_classification} presents the standard range of vehicle lengths and the number of axles proposed by the Transport Planning and Research Institute (TPRI)~\cite{TPRI}, Ministry of Transport, China. To identify the length of a vehicle using magnetic signals, one usually needs to first predict the speed of the vehicle, which could be estimated using the time-shift between the signals collected by the two sensors and the distance between the two sensors~\cite{liu2019vehicle}. With the speed predicted, the length of the vehicle could be estimated through obtaining the time spent by the vehicle for driving through the sensor node from nose to tail. 

With the vehicle length estimated, according to Table~\ref{table:vehicle_classification}, one could categorize vehicles into 4 types in a coarse-grained manner. To push the resolution of vehicle type classification, one could use supervised learning based on the collection of labeled signals. However, aforementioned solution is with significant limitations as follows: (1) From the raw magnetic signals illustrated in Figure~\ref{fig:proc_temp_freq}~(a), the exact time points that the vehicle arrives and departs from the sensor node are not obvious, and it is difficult to estimate the time spent by the vehicle for driving through the sensor node; (2) No matter from the temporal or frequency domains (shown in Figure~\ref{fig:proc_temp_freq}~(b)), the raw signals are less discriminability, and it is difficult to learn classifiers based on these signals. Thus, there needs methods to process the signals to \emph{shape the vibration of magnetic fields caused by the passing vehicles} and \emph{select discriminative features for vehicle type classification}.


\textbf{Contributions.} In our research, we propose \TheName, that intends to classify the vehicular type through magnetic sensing, signal processing, and semi-automated learning. Specifically, \TheName\ filters out noises and segments received magnetic signals by the exact time points that the vehicle arrives or departs from every sensor node. Further, \TheName\ adopts a hierarchical recognition model to first estimate the speed/velocity, and latter identify the length of vehicle using the predicted speed and the distance between the sensor nodes. With the vehicle length identified and the temporal/spectral features extracted from the magnetic signals, \TheName\ classify the types of vehicles accordingly. 

Compared to the general purpose magnetic approaches, with a collection of labeled signals, \TheName\ adopts several semi-automated learning techniques for the design of filters, features, and the choice of hyper-parameters via the comparison of cross-validation accuracy. Some examples of signals processed by \TheName\ (through interpolating, normalizing, and bandpass filtering) are illustrated in Figures~\ref{fig:proc_temp_freq}~(c) and~(d). In Figure~\ref{fig:proc_temp_freq}~(c), we could clearly observe the time duration that the vehicle drives through the sensor node from the processed signals in temporal domain. Further, Figure~\ref{fig:proc_temp_freq}~(d) shows the discriminability of processed signals in both temporal/frequency domains by the vehicle types. We have made at least following three contributions:

(1) We study the problem of vehicular traffic monitoring using magnetic field sensors, where we focus on vehicular type classification and traffic speed prediction using received magnetic field signals. To the best of our knowledge, it is the first work that aims at identifying the type of vehicles through measuring the speed and categorizing the vehicle lengths using a pair of magnetic sensors deployed on the straight section of highways, by tuning the performance of signal processing and recognition through semi-automated learning.


(2) We propose \TheName--- a magnetic sensing system with a hierarchical recognition model using a set of learning algorithms that can predict the traffic speed, identify the vehicle lengths, and classify the vehicular types accordingly. Specifically, \TheName\ adopts de-noising treatments to extract waveforms from raw signals, and match the signals collected by the two sensors through maximizing correlation coefficient between the two signals with a time-shift. With the time-shift estimated, \TheName\ predicts the speed/velocity of the vehicle with respect to the sampling rate of magnetic sensors and the distance between the two sensors. Later, \TheName\ filters out the noises from normalized signals in a bandpass fashion with respect to the predicted speed, and estimates the length of vehicle using a parameterizable white-box model. With a collection of labeled signals, \TheName\ tunes parameters of lowpass/highpass filters as well as the hyperparameters for length estimation through semi-automated learning via the comparisons of cross-validation accuracy. Finally, \TheName\ classifies the type of vehicle using the predicted length and Temporal/Frequency Domain Features extracted from signals. 


(3) We conduct extensive experiments based on real-world field deployment --- 4 \TheName\ systems deployed on the highways in Shenzhen, China. The experiment results show that \TheName\ can accurately identify the vehicular length and type; specifically, \TheName\ is with at least $90\%$ vehicle type classification accuracy with less than $5\%$ vehicle length classification error. \TheName\ can clearly outperform the existing systems~\cite{xu2017vehicle,xu2018vehicle,feng2020magmonitor} in both classification accuracy and the number of recognizable vehicle types (7 types by \TheName\ versus 4 types in~\cite{xu2017vehicle,xu2018vehicle,feng2020magmonitor}), while the deployment cost of \TheName\ is extremely low.

\section{Related Works \blue{and Discussion}} 
In this section, we review the relevant works in vehicle speed estimation, vehicle length estimation, and the vehicle types classification using magnetic sensors.

\blue{\subsection{Related Works}}
Magnetic sensing has been used in vehicular traffic surveillance~\cite{Caruso1999Vehicle,Cheung2007Traffic,zhang2010distributed,zhang2011real}, including detecting vehicles, estimating speed of vehicles, re-identify vehicle, and classifying the vehicle types. 
Incorporating Anisotropic Magneto-Resistive (AMR) signals measured from magnetic, \cite{Caruso1999Vehicle} propose to use AMR signals to classify vehicle types, including cars, vans, trucks, buses, trailer trucks, etc. The use of either single or multiple magnetic sensors has been studied in \cite{wahlstrom2014magnetometer}, where multi-sensor fusion for magnetic sensing has been used for vehicle type classification. Furthermore, various approaches have been studied to use magnetic sensing to classify the types of vehicles, such as~\cite{lan2009vehicle,jolevski2011smart,balid2017intelligent,balid2015development,gontarz2015use,bitar2016probabilistic,wang2017roadside,kleyko2015comparison}.

In addition to classification, vehicle tracking refers to monitoring the transient status of a moving vehicle. To achieve the goal, magnetic dipole models have been used ans studied in \cite{wahlstrom2014magnetometer,xie2014simulations,feng2020magmonitor,zhou2012practicable,gontarz2011impact,ren2015magnetic}, where variations of magnetic fields caused by the moving objects have been measured by magnetic sensors to track the vehicles. In addition to speed measurement, passing vehicle counting has been studied in~\cite{taghvaeeyan2013portable,Zhang2015A,wei2017adaptable,haoui2008wireless,obertov2014vehicle,li2011some,xiaoyong2010vehicle,feng2019magspeed,chen2019roadside,Kaewkamnerd2010Vehicle}, where the wave pattern matching of magnetic signals has been adopted in the algorithms. 
%

More specifically, \cite{Kaewkamnerd2010Vehicle} propose to use some advanced features that could be obtained in low-complexity, to achieve better classification accuracy for small vehicles. These features, including Average-Bar, Hillpattern peaks and magnetic signal differential energy, have been normalized to predict the vehicle speed and length. Furthermore, one could first measure the vehicle speed, then identify the vehicle length and use the vehicle length as the feature to classify vehicle types~\cite{de2016simple}. \blue{Novel millimeter-based Vehicle-to-Vehicle (VoV) and Vehicle-to-Infrastructure (VoI) communication techniques could be used to improve the overall performance of traffic monitoring~\cite{kong2017millimeter}}. Note that the performance evaluation and measurements in above work are all based on their own settings of deployments/experiments, which usually are not comparable with each other.



\blue{\subsection{Discussion and Comparisons to Our Work}}~
The most relevant studies to our work are~\cite{xu2017vehicle,velisavljevic2016wireless,feng2020magmonitor}. Particularly,~\cite{xu2017vehicle} proposed a portable roadside magnetic sensors for vehicle type classification, where they tried to classify vehicles into $4$ types with the classification accuracy $83.62\%$. Specifically, they use the Hill-Pattern, Peak-Peak, Mean-Std and Energy as features. Various supervised learners, including K-nearest neighbor (KNN), support vector machine (SVM) and back-propagation neural network (BPNN) have been used for classification. Velisavljevic's system \cite{velisavljevic2016wireless} could achieve classification accuracy rate of 88.37\% with 5 vehicle classes when was evaluated on a large realistic dataset. While all above works use common statistical learners, such as SVM, BPNN, KNN, K-means, perceptron neural network etc, using handcraft features extracted from either temporal/frequency domains, our work adopts automated machine learning \cite{nargesian2017learning,chen2019neural,kaul2017autolearn,kotthoff2017auto} that searches the empirically best classifiers with fine-tuned features and hyper-parameters to obtain the decent accuracy.

Despite the work~\cite{xu2017vehicle,velisavljevic2016wireless} achieves high recognition rate (88.37\% and 83.62\%), their method has the following limitations: 1) It does not solve the problem of noise on frequency waveform, and 2) the extracted features from the frequency waveform are not adequate to obtain all the distinctive properties of the vehicles. 
Generalization performance of existing classification work may be affected by too little realistic data. The proportion of vehicle type in existing work data set is far away to most country roads or urban road. 
Heuristically, speed and length of vehicle are those of the most important features that the magnetic sensor can detect precisely. Length can be used for classify some vehicles in big classification. Our works is the first use magnetic sensing measure vehicle speed and length for big classification, then use integral features of time-frequency domain get higher vehicle classification. Only use two sensor is another advantage of our solution. This design simple and robust. The most recent work, MagMonitor \cite{feng2020magmonitor}, introduces multiple magnetic dipole models. They propose to measure the magnetic signals from three orthogonal directions $(x, y, z)$ in the space, while our work only uses the magnetic signals in the $z$ direction while achieving better accuracy in recognition.

\begin{table}[!hbp]
\caption{Notations for \TheName}
\label{table:Notations}
\centering
\begin{tabular}{|l|l|}
\hline
Notations &  Meaning \\
\hline
$d$  & Two sensors distance \\
\hline
$f$  & ADC frequency of pre-processor \\
\hline
$\boldsymbol{x'}$ & Original signal vector from the first sensor \\
\hline
$\boldsymbol{x^{''}}$ & Original signal vector from the second sensor \\
\hline
$\boldsymbol{x}$ & Bandstop filterd signal vector for $\boldsymbol{x'}$\\
\hline
$\boldsymbol{x^{norm}}$ & Interpolated normalization for original signal vector \\
\hline
$\boldsymbol{x^{lh}}$ & Lowpass and highpass filtered signal vector for $\boldsymbol{x^{norm}}$\\
\hline
$\boldsymbol{X}$ & Frequency domain signal vector for $\boldsymbol{x^{norm}}$ \\
\hline
$\boldsymbol{X^{norm}}$ & Normalize frequency domain signal vector $\boldsymbol{X}$ \\
\hline
$\boldsymbol{X^{low}}$ & Low frequency of truncated vector $\boldsymbol{X^{norm}}$  \\
\hline
$\boldsymbol{F_s}$ & Chebyshev type I bandstop filter \\
\hline
$\boldsymbol{F_l}$ & Butterworth lowpass filter \\
\hline
$\boldsymbol{F_h}$ & Butterworth highpass filter \\
\hline
$\tau$ & Cross-correlation translation  \\
\hline
$v$  & Velocity of vehicle \\
\hline
$L^0$ & Real length of vehicle \\
\hline
$L$ & Length of vehicle \\
\hline
\end{tabular}
\end{table}

\section{\TheName \ System Design}
In this section, we first introduce the architectural design of \TheName\ with details of systems implementation. 

\subsection{Overall Architectural Design}
\begin{figure}
\centering
\includegraphics[scale=0.45]{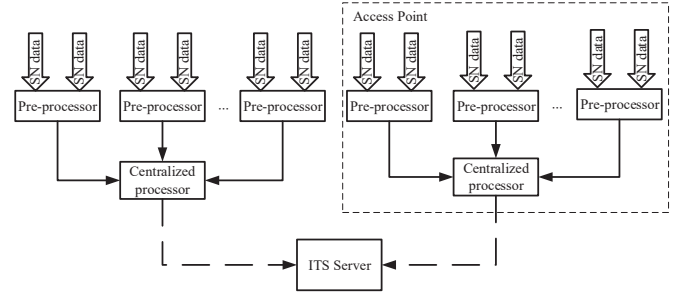}
\caption{Architectural Design of~\TheName~with a ITS Server.
}
\label{fig:architecture}
\end{figure}

Figure.~\ref{fig:architecture} illustrates the architectural design of \TheName. The purpose of \TheName\ is to provide real-time vehicular information to a central Intelligent Transportation Systems (ITS) server. More specifically, \TheName\ follows a Master-Slave design pattern with a hierarchical structure, where
\begin{itemize}
    \item A centralized processor is used to collect and aggregate sensing data from multiple parallel pre-processors and uploads the aggregated results to the central server; and
    \item Every pre-processor is connected to two magnetic sensors deployed on the road and the two magnetic sensors could capture the necessary magnetic signals for vehicle speeding detection, length estimation, and the type classification.
\end{itemize}
In this way, an ITS server could be deployed to monitor a large transportation system with road networks using multiple nodes (pre-processor and magnetic sensors).

\begin{figure}
\centering
\subfloat[Centralized Processor]{\includegraphics[width=0.32\textwidth, height=0.17\textheight]{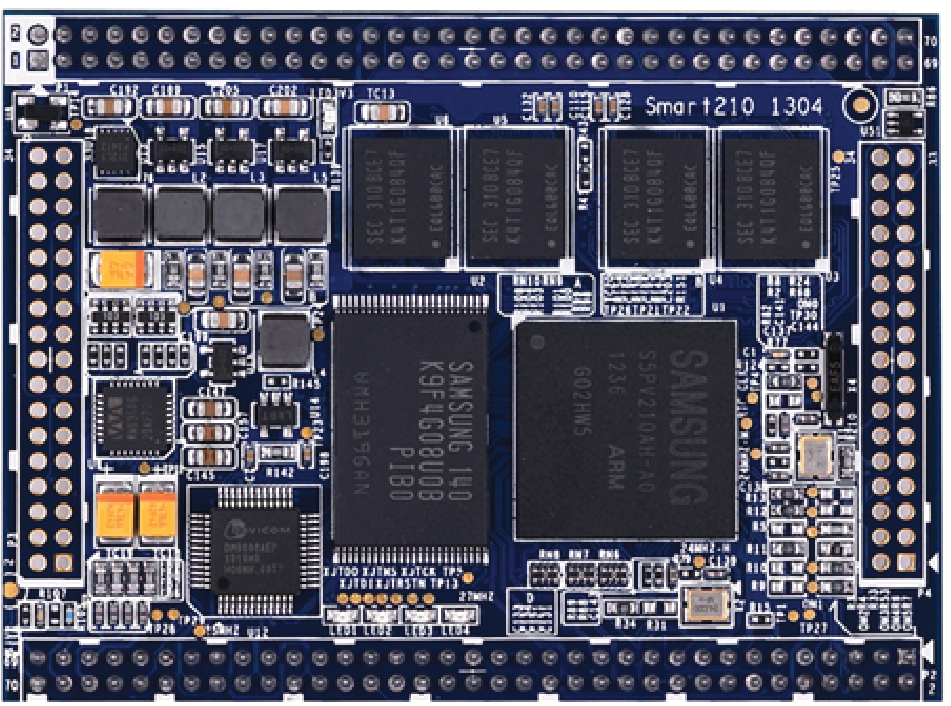}}\
\subfloat[Pre-processor]{\includegraphics[width=0.13\textwidth, height=0.07\textheight]{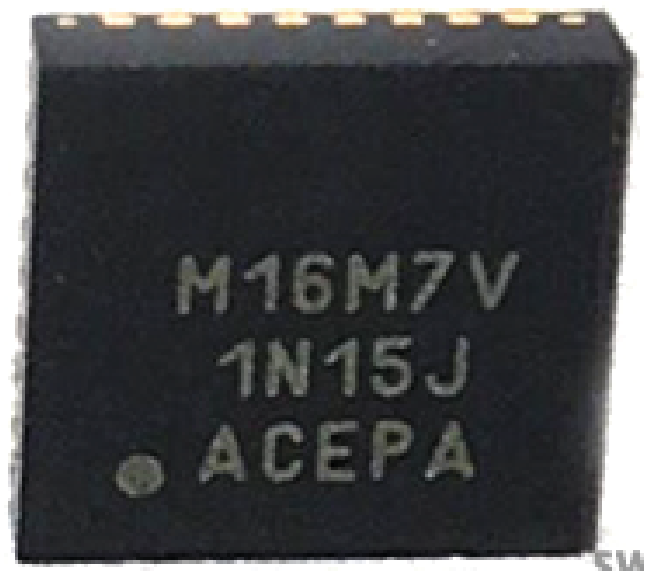}}\
\caption{Centralized Processor and Pre-processor .}
\label{fig:processor}
\end{figure}

\subsection{Design and Implementation of the Centralized Processor, Pre-processors, and the Paired Magnetic Sensors}
A centralized processor, as the master node of the system design, controls a set of pre-processors in an Access Point (AP)~\cite{balid2015development} shown in Figure~\ref{fig:architecture}. The centralized processor pulls every pre-processor in a round-robin manner and checks whether the pre-processor has any data to upload. Once the pair of magnetic sensors affiliated to the same pre-processor (through Analog-to-Digital Converter, ADC) detects signal vibrations for possible vehicle passing-by, they cache the signal data in the pre-processor and notifies the centralized processor for data transmission. Indeed, the pre-processor is in charge of providing the sensors reference voltages to control the electric circuit voltage in real-time. In addition, the pre-processor also adapts the sampling rate of ADC, so as to obtain the fine-grained time-series of magnetic signals while balancing the sensing cost. Note that one pre-processor with its affiliated sensor nodes could well cover an AP, which could refer to one section  of the road.

More specifically, with the magnetic signal data transmitted,  the centralized processor estimates the speed of vehicle through \emph{maximizing the correlation between the signals detected by the two sensors}. To avoid the potential affects of noises to the signal processing, \TheName\ proposes to use Bandstop filter to remove the interference, such as $50~\mathrm{Hz}$ alternating current (AC) frequency. 
With speed estimated, \TheName\ first refines the signals using Butterworth lowpass and highpass filters, then estimates vehicle length though regression. Based on the length of vehicle, the centralized processor categorizes vehicles into 4 types. Furthermore, the centralized processor extracts features in time-frequency domain from the signal and classifies those signal into 7 types by using automated machine learning. The final results, including location, time, vehicle number, speed, length, types etc. packaged by centralized processor, are sent to ITS server.

In our system design here, as shown in Figures~\ref{fig:processor}(a) and (b), the implementation of centralized processor is based on an Samsung S5PV210 single board computer with an ARM-11 CortexTM-A8 CPU (clock speed 1GHz), a main memory of 512MB DDR2 RAM (200Mhz), and a solid-state drive (SSD) disk for data storage. For every pre-processor, in this work, \TheName\ adopts a light-weighted  ARM Cortex M0+ 32-bit Microcontroller Unit (MCU) with a clock speed 48Hz, program storage 128KB, and data storage 16 KB, due to its low cost and low energy consumption. The resolution of Analog-to-Digital Converter (ADC) is 16 bit, which provides fine-grained magnetic signal readings over time. All these settings are adjustable to the realistic deployment of the systems.

\begin{figure}
\centering
\includegraphics[width=0.5\textwidth]{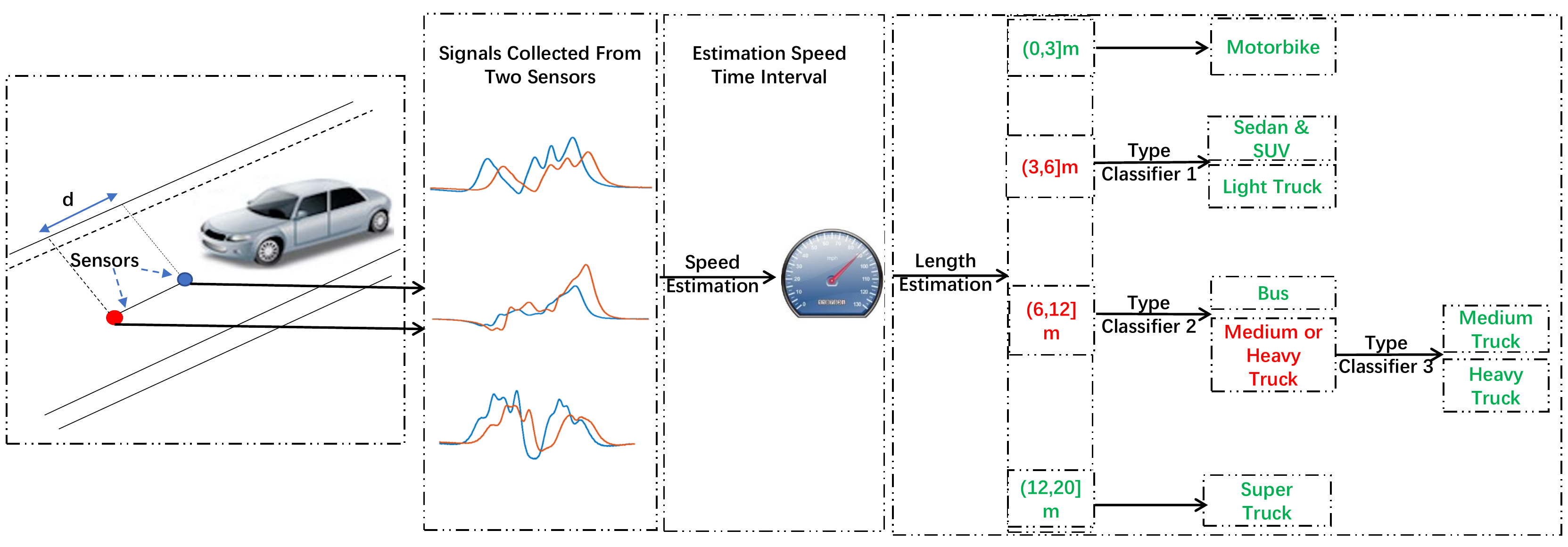}
\caption{The Hierarchical Recognition Model for Vehicle Type Classification.}
\label{fig:decisionTree}
\end{figure}

\section{\TheName \ Core Algorithms and Analysis}

Here, we  present design of the key \TheName\ algorithms. With the raw signal received, \TheName\ adopts a hierarchical recognition model demonstrated in Figure~\ref{fig:decisionTree}, where the proposed algorithms first predict the speed of vehicle, then identify the length of the vehicle with respect to the speed and the distance between sensor nodes, finally \TheName\ classifies the vehicle type using the estimated length and signal features.  



\begin{figure}
\centering
\subfloat[Original Signal]{\includegraphics[width=0.18\textwidth]{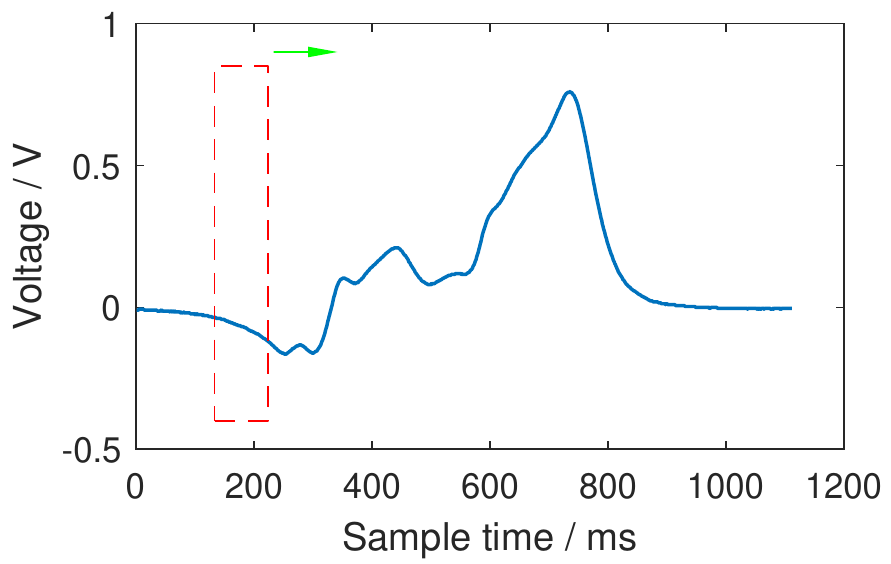}}
\subfloat[Standard Deviation]{\includegraphics[width=0.16\textwidth]{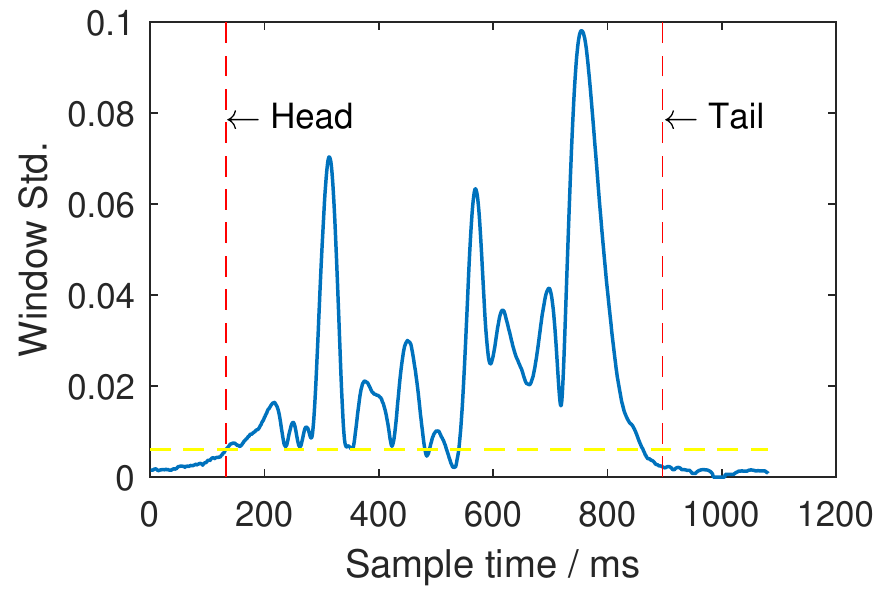}}
\subfloat[Extracted Waveform]{\includegraphics[width=0.16\textwidth]{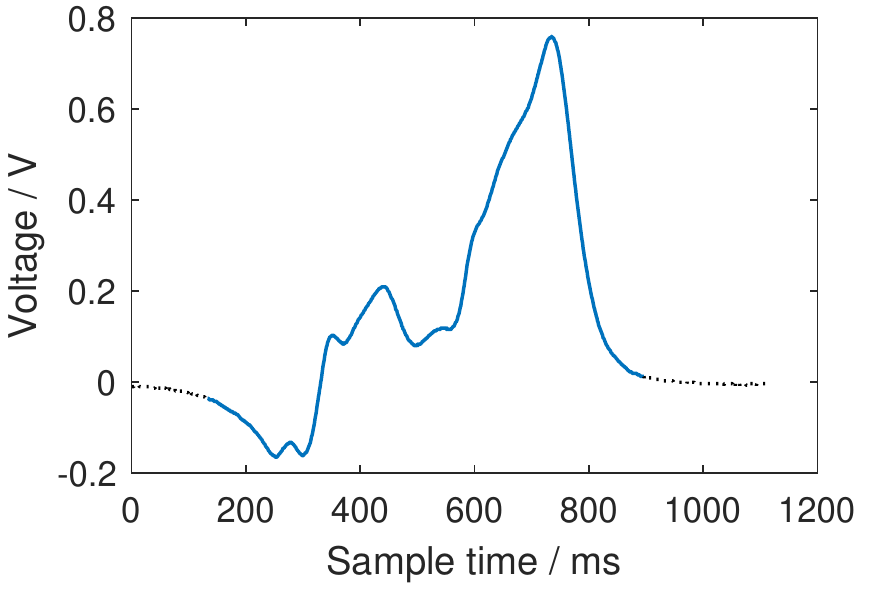}}\
\caption{Waveform Extraction using Standard Deviation and Sliding Window.}
\label{fig:orig_to_std}
\end{figure}

\begin{figure*}
\centering
\subfloat[Sedan and SUV]{\includegraphics[width=0.16\textwidth]{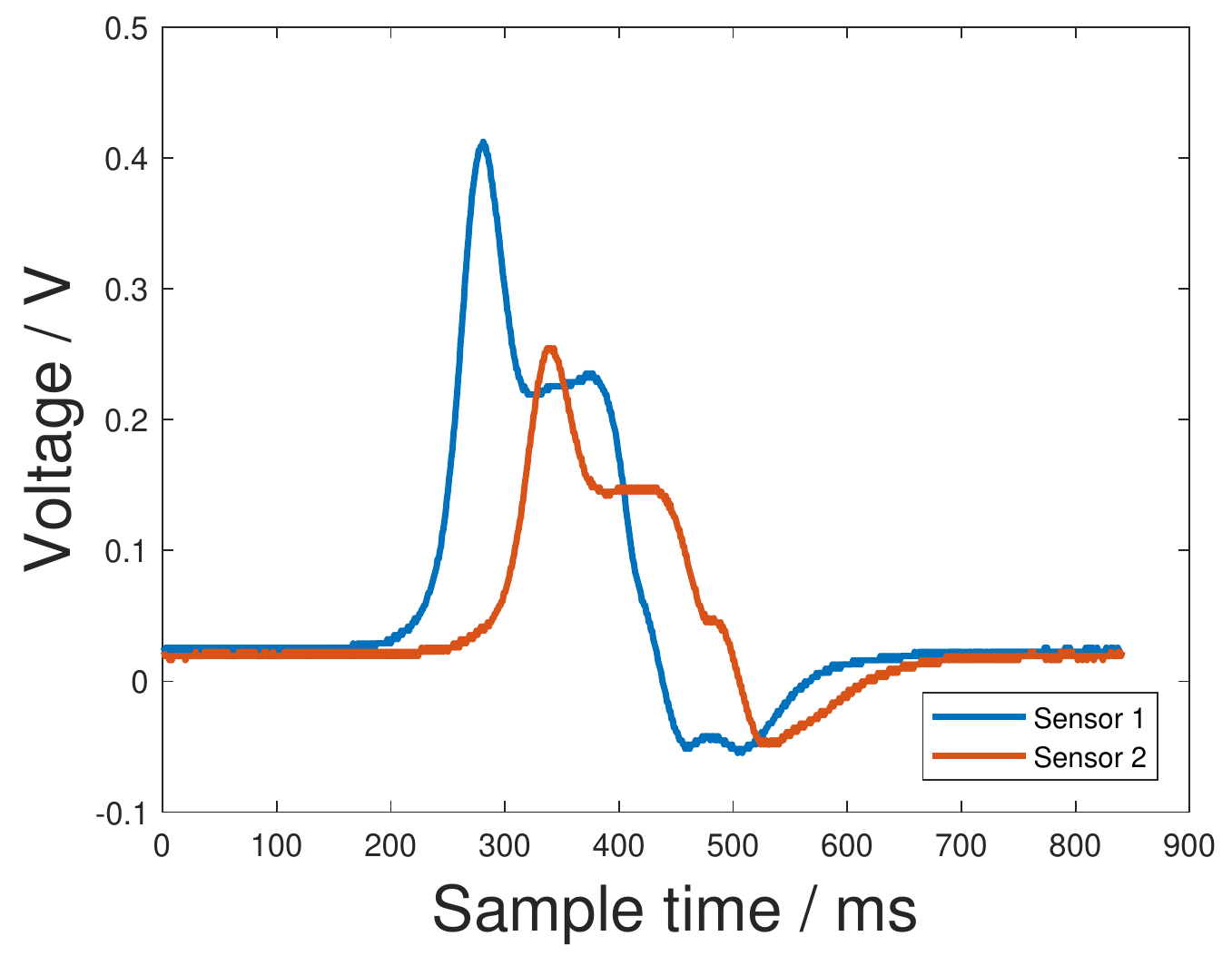}}\
\subfloat[Light Truck]{\includegraphics[width=0.16\textwidth]{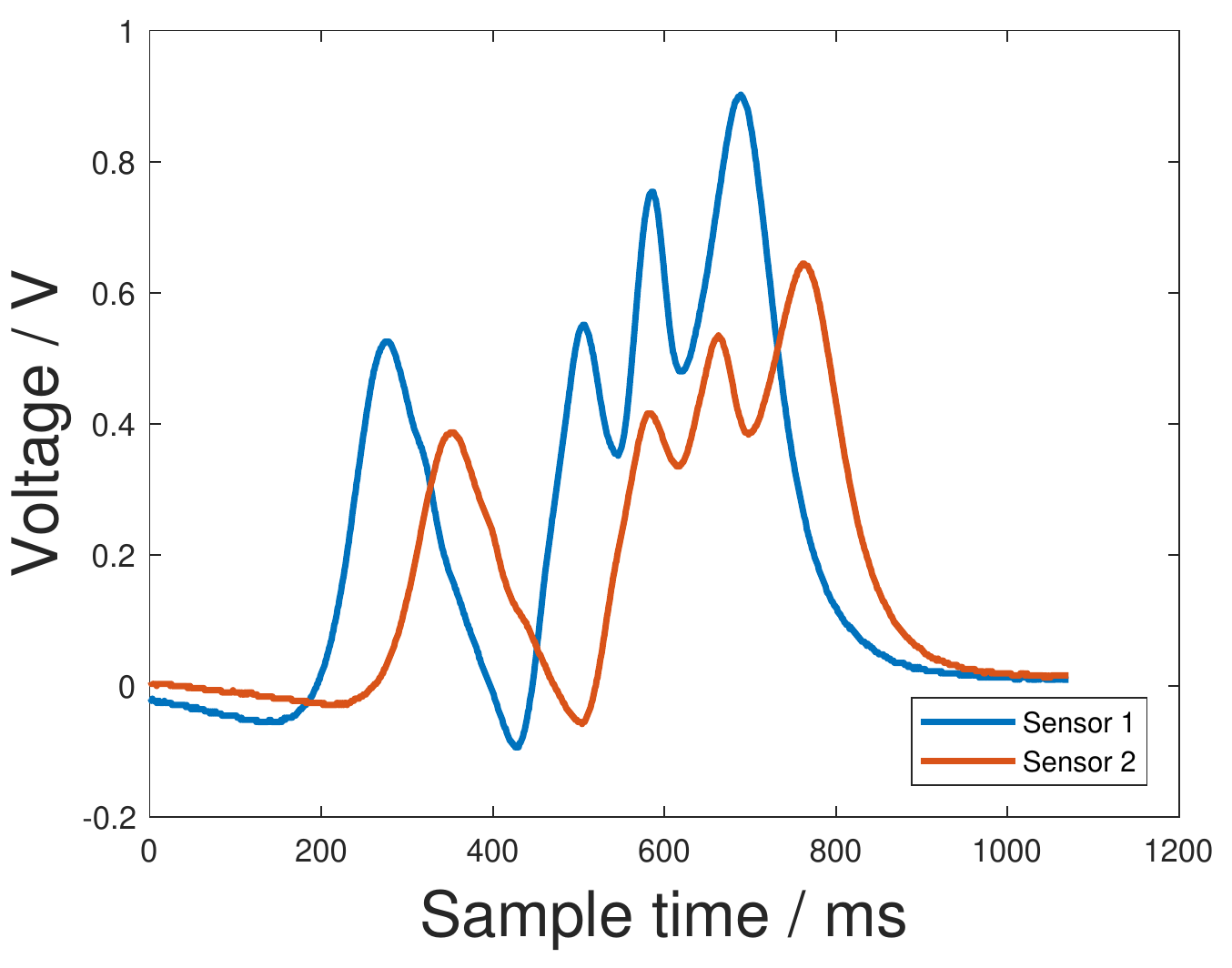}}\
\subfloat[Bus]{\includegraphics[width=0.16\textwidth]{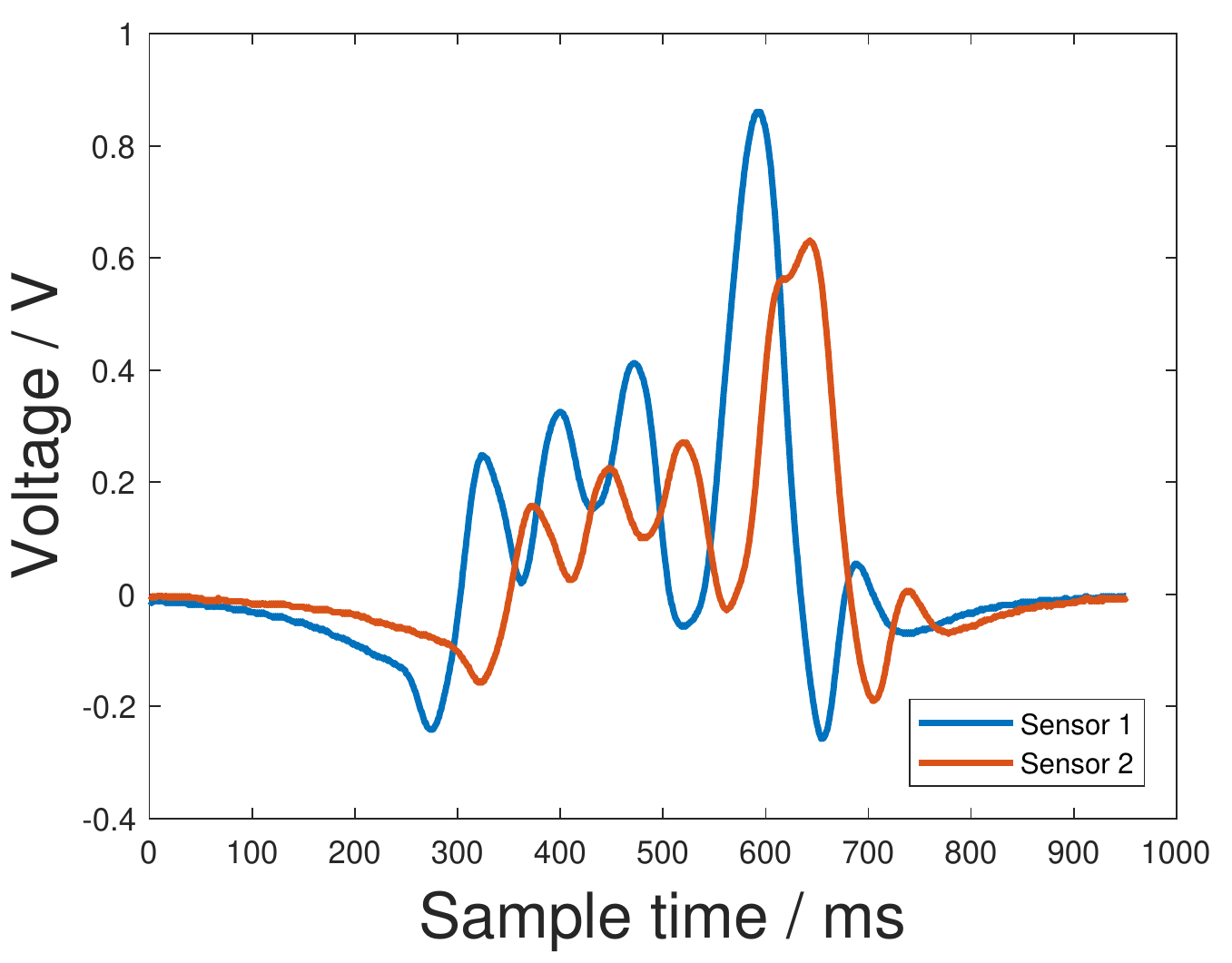}}\
\subfloat[Medium Truck]{\includegraphics[width=0.16\textwidth]{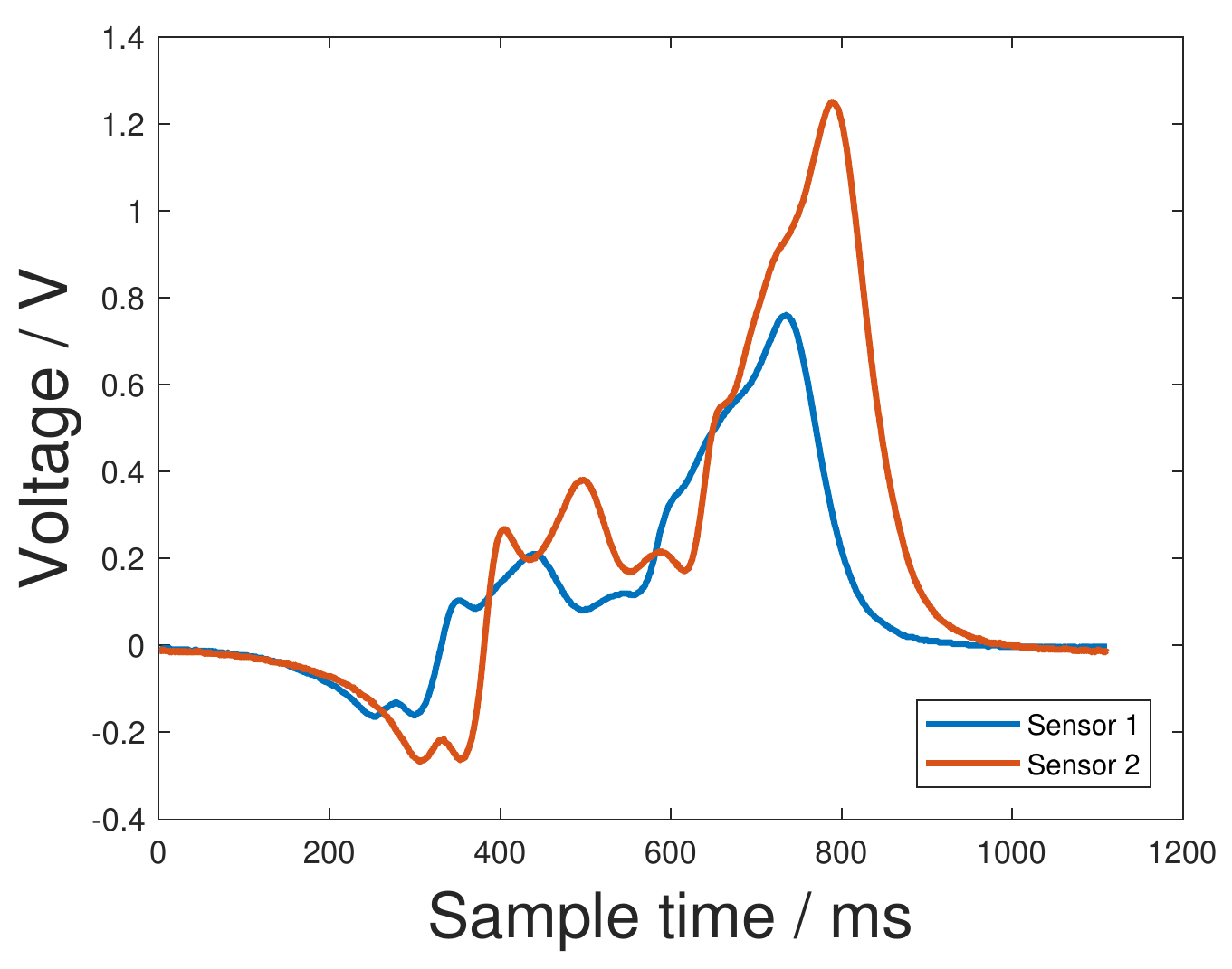}}\
\subfloat[Heavy Truck]{\includegraphics[width=0.16\textwidth]{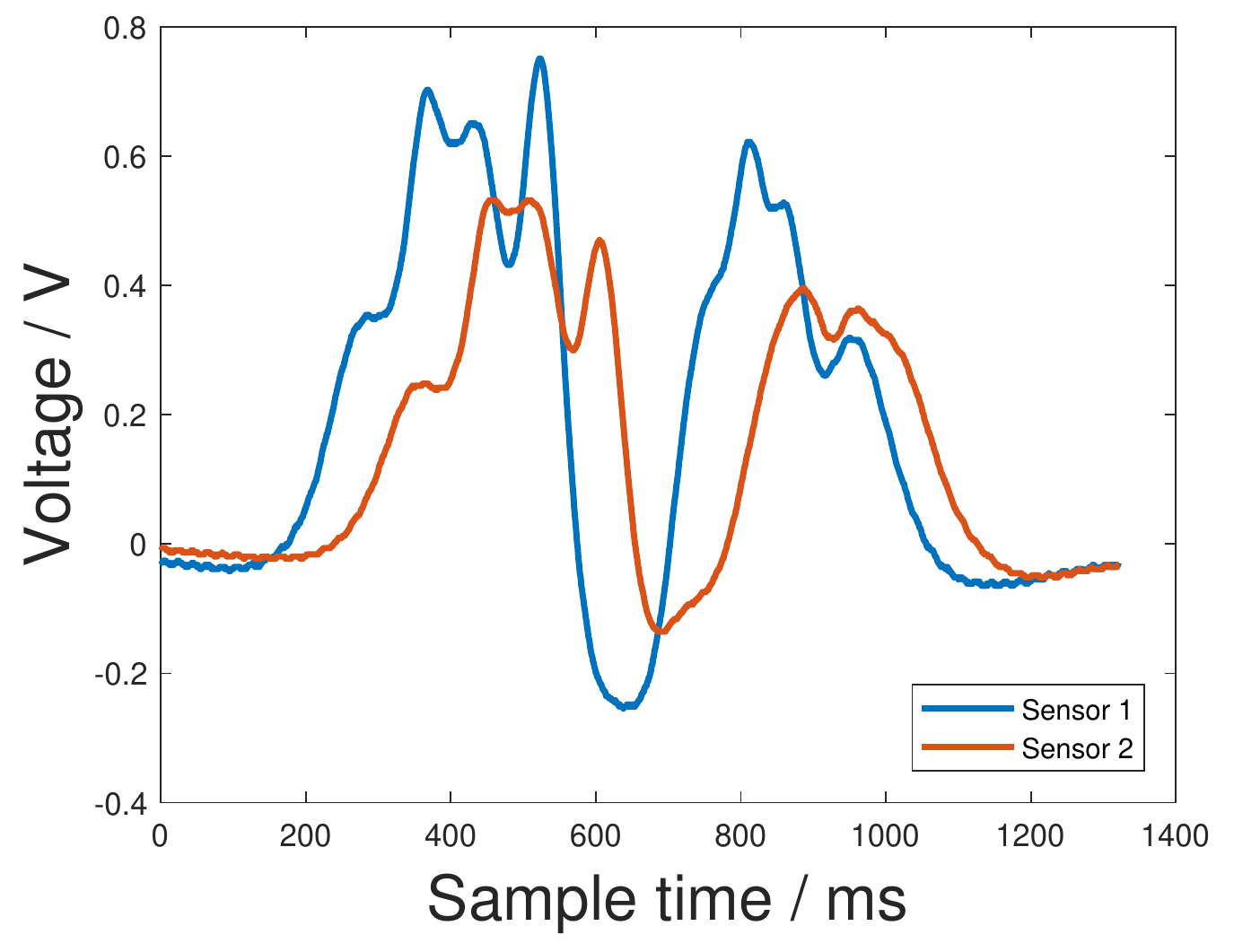}}\
\subfloat[Super Truck]{\includegraphics[width=0.16\textwidth]{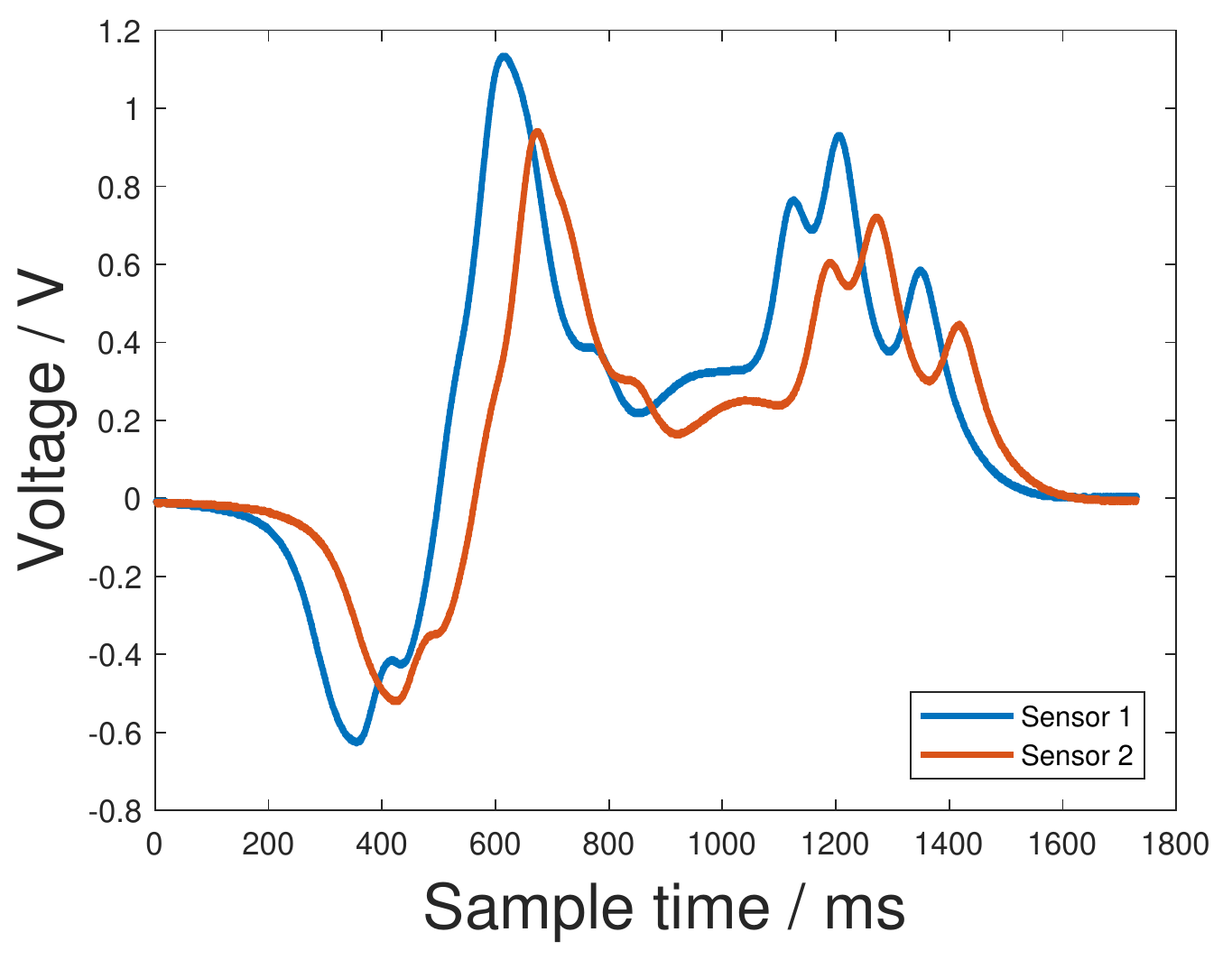}}\
\caption{Examples of Magnetic Signals Collected using Two sensors}
\label{fig:RawWaves}
\end{figure*}

\subsection{Vehicle Speed Estimation Models}
Given the magnetic signals collected from two sensors, this algorithm estimates the speed of vehicle for the further estimation of vehicle lengths and type classification. The algorithm consists of three steps: 
%
\begin{itemize}
    \item \textbf{Arrival/Departure Detection.} To estimate the speed, we first need to detect the vehicle that arrives and departs from a sensor node deployed on the road surface. With the raw ADC outputs, \TheName\ uses a sliding window to filter the time series of ADC outputs, and estimates the real-time standard deviation of ADC outputs in the sliding window. When the standard deviation in a time point is higher than the pre-defined threshold, \TheName\ identifies a significant change of signal in the time point and detects arrivals/departures of the vehicles to/from a sensor node.
    
    \item \textbf{Waveform Extraction.} Given the ADC outputs collected from one sensor node and the arrival/departure events detected, \TheName\ segments the ADC outputs by the events detected and extracts the magnetic signal change caused by the vehicles passing-by. Eventually, when a vehicle passes by, the magnetic sensor readings, as the ADC outputs, would achieve high levels and \TheName\ picks up the corresponding segments of ADC waveform as the signal caused by the vehicle passing-by.
    
    \item \textbf{Waveforms Matching.} Given waveforms extracted from two paired sensors, \TheName\ estimates the speed of vehicle through matching the two waveforms. Figure~\ref{fig:RawWaves} illustrates the example of magnetic signals collected from two sensor nodes deployed in the same road section, where we can observe a clear time-shift as the vehicle passes the two sensor nodes sequentially with a time gap. Furthermore, as the two waveforms are both caused by the same passing vehicle, they should share a similar pattern in the time domain. One could detect the speed/velocity of the vehicle (denoted as $v$) through calculating the time shift $\hat{\tau}$ of two waveforms. We define the sampling frequency of ADC is $f$ Hz and the distance between two sensors is $d$ meter. Specifically, we estimate the velocity $v$ km/h as follow
    \begin{equation}
        v= 3.6 \ \frac {d\cdot f}{\hat{\tau}} \ \mathrm{km/h}\ ,
    \end{equation}
    where the time shift $\hat\tau$ is estimated through maximizing the correlation between the time-shifted waveforms
    \begin{equation}
    \begin{split}
    \label{equ:v}
    &\hat{\tau} = \underset{\tau \in (0, 200)}{\mathrm{argmax~coef(\tau)}}, \ \text{where} \\ 
    &\mathrm{coef}(\tau)=\frac{\sum_{i=1}^{n}(\boldsymbol{x}_{i}' -\bar{x'})(\boldsymbol{x}_{i+\tau}^{''}-\bar{x^{''}})} {n\sqrt{\sum_{i=1}^n(\boldsymbol{x}_{i}'-\bar{x'})^2}\sqrt{\sum_{i=1}^n(\boldsymbol{x}_{i}^{''}-\bar{x^{''}})^2}}\ .
    \end{split}
    \end{equation}
\end{itemize}
Note that $n$ refers to the sample points in the waveforms, $\boldsymbol{x'}$ and $\boldsymbol{x^{''}}$ refer to the two waveforms (represented as two $n$-dimensional vectors) respectively, $\bar{x'}$ is the mean in $\boldsymbol{x'}$, $\bar{x^{''}}$ is the mean in $\boldsymbol{x^{''}}$, $\tau \in (0,200)$ refers to the potential time shift as the velocity $v \in (0,200)$ km/h, $d=1$ meter and $f=1000$Hz in our experiment.

\subsection{Vehicle Length Estimation Models}
Given the estimated speed and the two waveforms, \TheName\ predict the length of vehicle through classifying the length of vehicle. Specifically, we categorize the length in four sets $(0,3]m$, $(3,6]m$, $(6,12]m$ and $(12,20]m$, which corresponds to the standard ranges of vehicle lengths proposed by the Transport Planning and Research Institute (TPRI)~\cite{TPRI}, Ministry of Transport, China.

\begin{algorithm}
  \caption{Semi-Automated Learning}
  \label{alg:Vehicle_Length}
  \begin{algorithmic}[1]
    \Require
      Datasets of the signal waveforms $\boldsymbol{X'}$;
      Minimum velocity of a vehicle is $v_{min}$;
      Estimated Speed $v$;
    \Ensure Lowpass/Highpass Filters $\boldsymbol{F_l}$  and $\boldsymbol{F_h}$; Proportion of Fade-in/Fade-out Periods $c$ in the signal.
   \State $\boldsymbol{X}^\mathrm{norm}\gets\emptyset$\ \textcolor{blue}{/*Initialization*/}
   \For{$\forall \boldsymbol{x'}\in \boldsymbol{X'}$}
        \State $\boldsymbol{x}^\mathrm{norm}\gets\mathrm{interpolate}(\boldsymbol{x}')$ \ \textcolor{blue}{/*Using Eq.~\ref{equ:interpolation}*/}
        \State $\boldsymbol{X}^\mathrm{norm}\gets \boldsymbol{X}^\mathrm{norm}\cup \{\boldsymbol{x}^\mathrm{norm}\}$\ \textcolor{blue}{/*Adding to $\boldsymbol{X}^\mathrm{norm}$*/}
    \EndFor
    \Repeat
        \State Search the parameters for filters $\boldsymbol{F_l}$ and $\boldsymbol{F_h}$
        \State Search the proportion of fade-in/fade-out periods $c$
        \State $\mathrm{Error}\gets 0$
        \For{$\forall \boldsymbol{x}^\mathrm{norm}\in \boldsymbol{X}^\mathrm{norm}$}
        \State $\boldsymbol{x^{lh}}\gets\boldsymbol{F_l}(\boldsymbol{F_h}(\boldsymbol{x}^\mathrm{norm}))$\ \textcolor{blue}{/*Bandpass Filtering */}
        \State $L\gets\mathrm{Length Estimator}(\boldsymbol{x^{lh}},c)$\ \textcolor{blue}{/*Using Eq.~\ref{equ:vehiclelength}*/}
        \If{$L\not\in$ range of length by the type}
            $\mathrm{Error}++$
        \EndIf
        \EndFor
    \Until{find the minimum $\mathrm{Error}$}.  \\
    \Return Length types.
  \end{algorithmic}
\end{algorithm}


{\bf Off-line training.} To achieve the goal, \TheName\ adopts automated learning techniques to train the classifier for the prediction. Specifically, there consists of two steps as follow.
\begin{itemize}
    \item \textbf{Data Collection.}~Given waveform extracted from both sensor nodes, we first try to build up a training dataset for vehicle length estimation with a given set of vehicles as examples. As what would be disclosed in Section V, we totally include $1,000$ vehicles in the data collection, where examples of Motorbike (15), Sedan and SUV (350), Light Truck (280), Medium Truck (120), Heavy Truck (85), Super Truck (50), and Bus (100) are included. With waveforms collected for these $1,000$ vehicles from the real-world deployments of sensor nodes in xxx road sections, we label every piece of waveform with their corresponding vehicle types and vehicle length.

    \item \textbf{Semi-Automated Learning.}~Given collected datasets $\boldsymbol{X}'$, the goal of this step is to optimize the parameters of a bandpass filter that processes original magnetic signal data from the frequency domain and extracts discriminative features for vehicle length/type classification (See examples in Fig.~\ref{fig:proc_temp_freq}). More specifically, we propose using the semi-automated feature extraction algorithm listed in Algorithm~\ref{alg:Vehicle_Length} to search the parameters for the filters. First of all, given every sample in the signal dataset $\boldsymbol{x}'\in\boldsymbol{X}'$, \TheName\ interpolates/normalizes the signal and obtains the $\boldsymbol{x}^\mathrm{norm}$ using the moving smoother as follow.
    \begin{equation}
        \begin{split}
        \label{equ:interpolation}
        \boldsymbol{x_{i}}^\mathrm{norm}=(1-\frac{v_{min}}{v})\ \boldsymbol{x'_{i}}+\frac{v_{min}}{v}\ \boldsymbol{x'_{i+1}} 
        \end{split}
    \end{equation}
    where $\boldsymbol{x}_i^\mathrm{norm}$ refers to the $i^{th}$ point in the discrete-time signal ($n$ points in total supposed), $d$ refers to the distance between two sensor nodes, and $v_{min}$ refers to the minimal velocity of vehicles. Given the interpolated and normalized datasets $\boldsymbol{X}^\mathrm{norm}$, \TheName\ uses a pair of butterworth lowpass/highpass filters (See examples in Fig.~\ref{fig:band_pass_filter}), denoted as $\boldsymbol{F_l}$ and $\boldsymbol{F_h}$ respectively, for bandpass filtering. The algorithm tunes the parameters of the filters. The objective of parameter search is to minimize the training error of length classification using the filtered data $\boldsymbol{x^{lh}}=\boldsymbol{F_l}(\boldsymbol{F_h}(\boldsymbol{x}^\mathrm{norm}))$ for $\forall\boldsymbol{x}^\mathrm{norm}\in \boldsymbol{X}^\mathrm{norm}$.  Specifically, \TheName\ estimates the length of vehicle $L$ (in meters) as follow,
    \begin{equation}
    \label{equ:vehiclelength}
    L=\frac{d\cdot \mathrm{Cyc}(\boldsymbol{x^{lh}},c)} {\hat{\tau}\cdot v } = \frac{\mathrm{Cyc}(\boldsymbol{x^{lh}},c)}{3.6\cdot f} \ .
    \end{equation}
    where $f$ refers to the sampling frequency of ADC. $\mathrm{Cyc}(\boldsymbol{x^{lh}},c)$ counts the effective number of ADC sensing cycles in the filtered signal $\boldsymbol{x^{lh}}$ under the tuning parameter $c$, which refers to the time spent by the vehicle to pass by the sensor mode. To count the number of effective ADC sensing cycles, \TheName\ estimates the area under curve of $\boldsymbol{x^{lh}}$ (i.e., total energy) in temporal domain and removes the fade-in and fade-out periods in the signal, where the fade-in and fade-out periods are supposed to take $c$ proportion (e.g., several percentage) of the area under curve (total energy). Note that the choice of $c$ is also a part of parameter search and the optimal setting searched in our study is $c=4\%$.

    \begin{figure}[!ht]
    \centering
    \subfloat[Butterworth Lowpass Filter]{\includegraphics[width=0.23\textwidth, height=0.12\textheight]{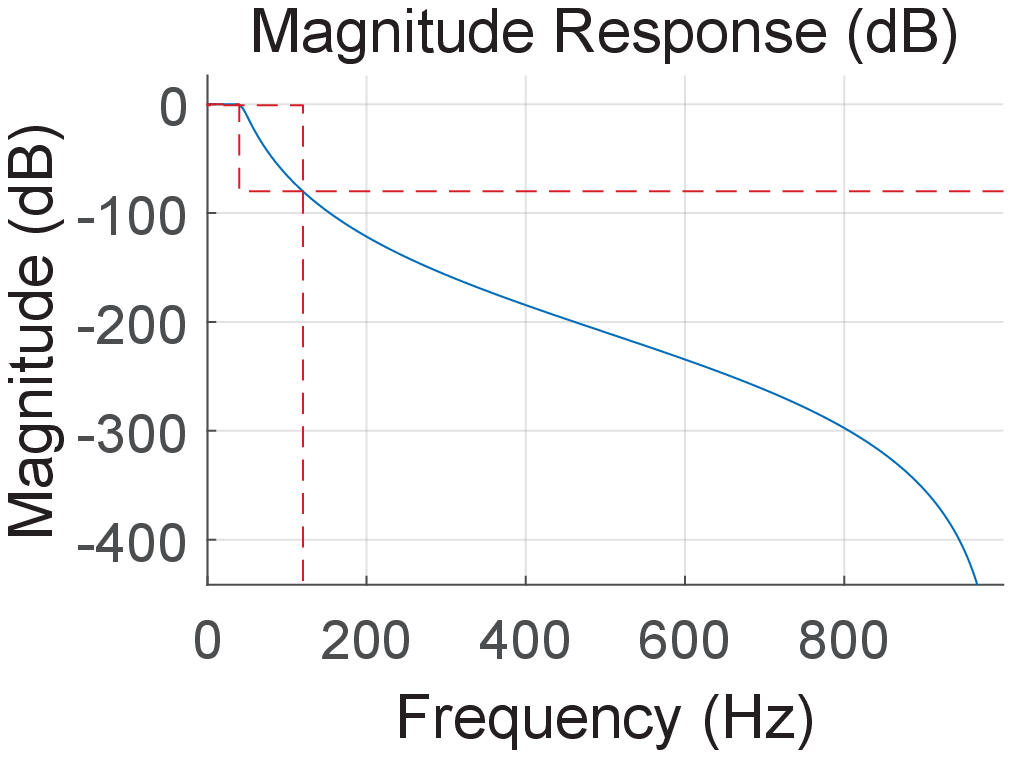}}\
    \subfloat[Butterworth Highpass Filter]{\includegraphics[width=0.23\textwidth, height=0.12\textheight]{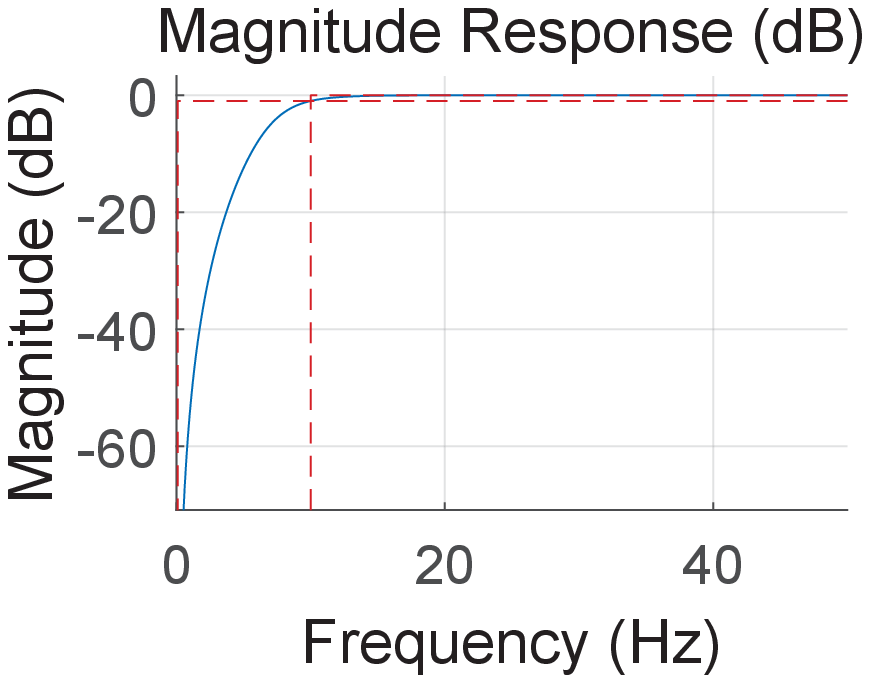}}\
    \caption{Butterworth Lowpass and Highpass Filters. Filters cut the low and high energy of vehicle waveform. The remaining frequency bands are relatively stable in energy.}
    \label{fig:band_pass_filter}
    \end{figure}
    
%


    
    

\end{itemize}



{\bf On-line prediction.} Given a new sample of signal obtained in on-line setting, \TheName\ first uses lowpass/highpass filters to process the signal, then uses the estimator in Eq.~\ref{equ:vehiclelength} for length estimation, and categorizes the length in four sets $(0,3]m$, $(3,6]m$, $(6,12]m$ and $(12,20]m$. Finally, \TheName\ forwards the results of length estimation and the processed signal for the further vehicle type classification.

\subsection{Vehicle Type Classification Models}
Given the estimated vehicle length, the sample of interpolated and normalized signal $\boldsymbol{x}^\mathrm{norm}$, \TheName\ employs a hierarchical classification models (See also in Figure~\ref{fig:decisionTree}) to identify the types of vehicle in fine-grained, i.e., from 4 categories of vehicle length to 7 types of vehicles. Specifically, \TheName\ extracts additional 8 features from $\boldsymbol{x}^\mathrm{norm}$ to train 3 classifiers to make binary classification: (1) between Sedan vs Light Truck, (2) between Bus vs Median Truck or Heavy Truck, and (3) between Median Truck vs Heavy Truck respectively.


The extracted features are defined in Table~\ref{table:Features}. The features in temporal domains are calculated as follow.
    \begin{equation}
    \label{equ:COG}
    c_t=\frac{\sum_{i=1}^{n}i\cdot |\boldsymbol{x}_{i}^\mathrm{norm}|}{\sum_{i=1}^{n}|\boldsymbol{x}_{i}^\mathrm{norm}|}
    \end{equation}

    \begin{equation}
    \label{equ:dispersion}
    d_t=\frac{\sum_{i=1}^{n}(i-c_t)^2\cdot |\boldsymbol{x}_{i}^\mathrm{norm}|}{\sum_{i=1}^{n}|\boldsymbol{x}_{i}^\mathrm{norm}|}
    \end{equation}
Then \TheName\ carries out Fast Fourier Transformation (FFT) to obtain the spectral information as follow.
    \begin{equation}
    \label{equ:fft}
    \begin{split}
    \boldsymbol{s} = \mathrm{FFT}(\boldsymbol{x}^\mathrm{norm}) \\
    \boldsymbol{s}_{i}^\mathrm{norm} = \boldsymbol{s}_{i} \cdot \frac {f}{n}  \ \ \forall i \in [1,n] \\
    \boldsymbol{s}_{i}^\mathrm{low} = \boldsymbol{s}_{i}^\mathrm{norm}   \ \  \forall i \in [1:20]
    \end{split}
    \end{equation}
    Note that the interpolation and normalization makes the frequency resolution as 1Hz, and we only consider the spectral data below 20 Hz as the low band spectral features.

    \begin{table}
    \caption{Features Extracted From Vehicle Signal}
    \label{table:Features}
    \centering
    \begin{tabular}{|l|l|}
    \hline
    Features &  Meaning \\
    \hline
    $L$& vehicle length, equation (\ref{equ:vehiclelength}) \\
    \hline
    $\bar{x_t}$  &  Average signal strength in time domain \\
    \hline
    $\sigma_t$  & MSE in time domain \\
    \hline
    $c_t$  &  Center of gravity in time domain, equation (\ref{equ:COG}) \\
    \hline
    $d_t$  & Dispersion in time domain, equation (\ref{equ:dispersion}) \\
    \hline
    $\bar{x}_f$ & Average amplitude in frequency domain \\
    \hline
    $\sigma_f$ & MSE in frequency domain \\
    \hline
    $c_f$  & Center of gravity in frequency domain, equation (\ref{equ:COG_f}) \\
    \hline
    $d_f$  & Dispersion in frequency domain, equation (\ref{equ:dispersion_f}) \\
    \hline
    \end{tabular}
    \end{table}

With the spectral information, \TheName\ forms the features in frequency domains as follow.
    \begin{equation}
    \label{equ:COG_f}
    c_f=\frac{\sum_{i=1}^{n}i\cdot |\boldsymbol{s}_{i}^\mathrm{low}|}{\sum_{i=1}^{n}|\boldsymbol{s}_{i}^\mathrm{low}|}
    \end{equation}

    \begin{equation}
    \label{equ:dispersion_f}
    d_f=\frac{\sum_{i=1}^{n}(i-c_f)^2\cdot |\boldsymbol{s}_{i}^\mathrm{low}|}{\sum_{i=1}^{n}|\boldsymbol{s}_{i}^\mathrm{low}|}
    \end{equation}
    
    
    \begin{figure}[!ht]
    \centering
    \subfloat[Original Waveform]{\includegraphics[width=0.23\textwidth]{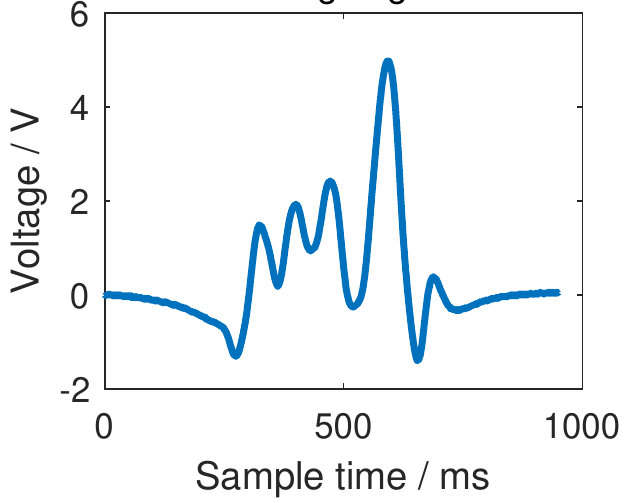}}\
    \subfloat[Interpolated Waveform]{\includegraphics[width=0.23\textwidth]{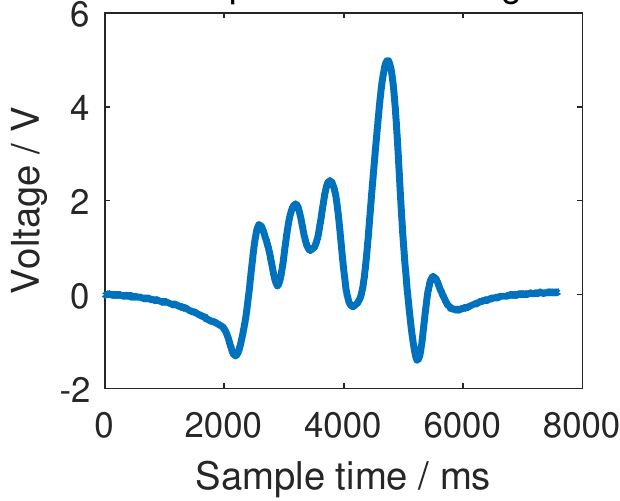}}\
    \subfloat[FFT Spectrum]{\includegraphics[width=0.23\textwidth]{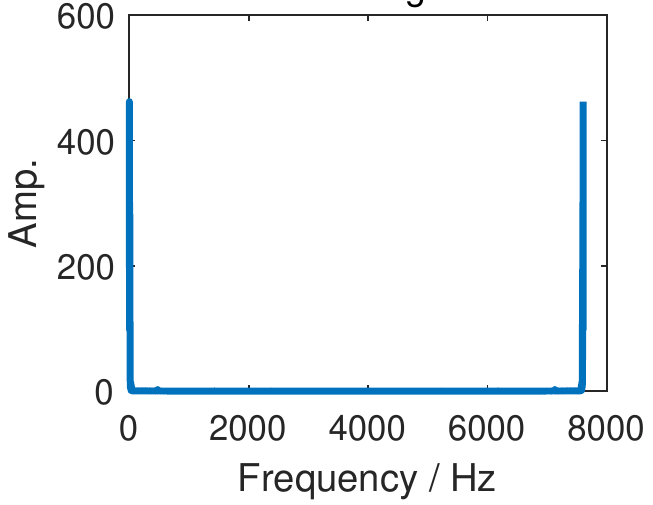}}\
    \subfloat[Lowband Spectrum]{\includegraphics[width=0.23\textwidth]{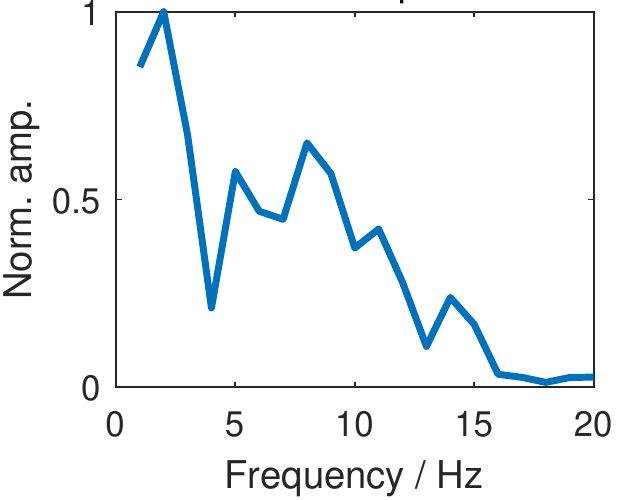}}\
    \caption{Examples of Original Signals, Interpolated Signals with Refined Resolution, Full Spectrum by FFT, and the Low Band Spectrum.}
    \label{fig:orig_to_truncate}
    \end{figure}
An example of original signals, interpolated signals, full spectral data by FFT, and the low band spectrum is given in Figure~\ref{fig:orig_to_truncate}. The low band spectrum well characterizes the vibration of magnetic signals caused by the approaching heavy vehicles.  After all, we use all these features to learn the three classifiers using the collected datasets and Support Vector Machine \blue{with Radical Basis Function Kernels (RBF-Kernel SVM)~\cite{chapelle1999support}}, where the 9 original features are used.

\begin{figure}
\centering
\includegraphics[width=0.48\textwidth]{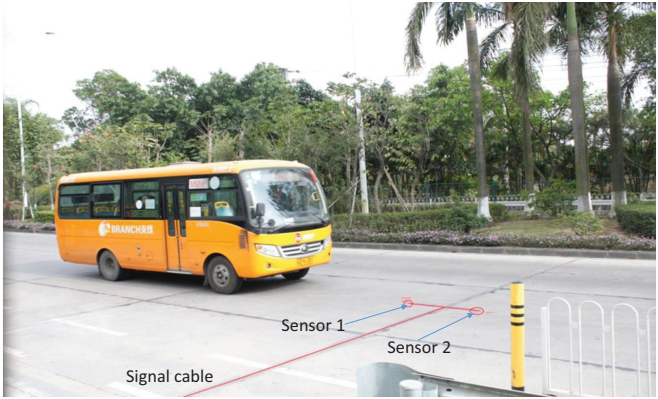}
\caption{Examples of Real-world Deployment: Two magnetic sensors were deployed in the road; the distance between the two sensors is 1 meter; and the collected data were sent to pre-processors and centralized processor by cables.}
\label{fig:expSet}
\end{figure}

\section{Experiments}

In this section, we verify the performance of the proposed method with the real-world deployment. Specifically, we first present the experimental settings, and then demonstrate the experimental results and further analysis.

\subsection{Experimental Settings}
The data, including $7$ types of vehicles, utilized in the experiments were collected from the roads in Shenzhen, China (see Fig. 9). In order to prevent over-fitting, we randomly select $1000$ data pairs for model training, and the rest $1153$ data pairs for model performance testing. The details of the testing set are listed in Table IV, in which the lengths of $15$ vehicles are $L\in(0,3] m$, $841$ vehicles are $L\in(3,6] m$, $273$ vehicles are $L\in(6,12] m$, and $24$ vehicles are $L\in(12,20] m$. 

To collect data for training and testing, we use a set of commonly-used filters to process signal data. Chebyshev type I bandstop filter is designed using FDESIGN.BANDSTOP. The sampling frequency is $f=1000$ Hz, order is 2, first passband frequency is 25 Hz, second passband frequency is 100 Hz, passband ripple is 1 dB. Butterworth highpass filter is designed using FDESIGN.HIGHPASS. Highpass filter sample frequency is 2000 Hz, stopband frequency is 0.1 Hz, passband frequency is 10 Hz, stopband attenuation is 80 dB and passband ripple 1 dB. Butterworth lowpass filter is designed using FDESIGN.LOWPASS. Lowpass filter sampling frequency is 2000 Hz, passband frequency is 40 Hz, stopband frequency 120 Hz, stopband attenuation 80 dB and passband ripple is 1 dB. With all data collected, all computation was conducted on a laptop with an AMD CPU, 8G RAM, and Windows 10 operating system, using MATLAB R2017b.

\begin{table}
\caption{Confusion Matrix of Vehicle Length Classification}
\label{table:4types}
\centering
\begin{tabular}{|c|c|c|c|c|}
\hline
\diagbox{Truth}{Predicted} & $(0,3] m$ & $(3,6] m$ & $(6,12] m$ & $(12,20] m $   \\
\hline
$(0,3] m$ & \textbf{15}  & \red{0} & \red{0}  & \red{0}     \\
\hline
$(3,6] m$ & \red{0} & \textbf{806}  & \red{35}  & \red{0}   \\
\hline
$(6,12] m$ & \red{0} & \red{16} & \textbf{255}  & \red{2}   \\
\hline
$(12,20] m$ & \red{0} & \red{0} & \red{1} & \textbf{23}   \\
\hline
\end{tabular}
\end{table}

\begin{table*}
\caption{Confusion Matrix of Vehicle Type Classification using \TheName.}
\label{table:7types}
\centering
\begin{tabular}{|c|c|c|c|c|c|c|c|}
\hline
\diagbox{Truth}{Predicted}  & Motorbike & Sedan and SUV & Light Truck & Bus & Medium Truck & Heavy Truck & Super Truck \\
\hline
Motorbike & \textbf{15}  & \red{0} & \red{0}  & \red{0}  & \red{0} & \red{0} & \red{0} \\
\hline
Sedan and SUV & \red{0} & \textbf{499}  & \red{12}  & \red{12}  & \red{8} & \red{1} & \red{0} \\
\hline
Light Truck & \red{0} & \red{8} & \textbf{287}  & \red{5} & \red{3} & \red{6} & \red{0} \\
\hline
Bus & \red{0} & \red{2} & \red{2} & \textbf{66}  & \red{3} & \red{3} & \red{0} \\
\hline
Medium Truck & \red{0} & \red{2} & \red{5} & \red{9} & \textbf{125} & \red{19} & \red{1} \\
\hline
Heavy Truck & \red{0} & \red{0} & \red{5} & \red{2} & \red{2} & \textbf{26} & \red{1} \\
\hline
Super Truck & \red{0} & \red{0} & \red{0} & \red{0} & \red{0} & \red{1} & \textbf{23} \\
\hline
\end{tabular}
\end{table*}

\begin{table*}
\caption{Overall Comparison Results. N/A: details were not disclosed in the paper.}
\label{table:Solution_Comparison}
\centering
\begin{tabular}{c|c|l|l|l|c}
\hline
Solutions &  Train/Test Data & Features & Classifiers & Precision\%/Recall\% per Type & Overall Accuracy (\%) \\
\hline
Xu \emph{et al}.\cite{xu2017vehicle}& 140 / 37 &   \makecell*[l]{Hill-Pattern,\\ Peak-Peak,\\ Mean-Std, and\\ Energy all in\\ Temporal Domain} &  \makecell*[l]{KNN,\\ \blue{RBF-Kernel} SVM, and\\ Neural Networks} &  \makecell*[l]{Two-box (N/A),\\ Saloon (N/A),\\ Bus (N/A), and\\ MPV (N/A)} & 83.6 \\
\hline
SMOTE~\cite{xu2018vehicle}&  130 / 44  & \makecell*[l]{3D Magnetic\\ Signals, and FFT } & \makecell*[l]{KNN,\\ \blue{RBF-Kernel} SVM, and\\ Neural Networks}  &  \makecell*[l]{Hatchback (80.0/100.0),\\ Sedan (100.0/92.3),\\ Bus (100.0/100.0),\\ and MPV (75.0/100.0)} &  95.5 \\
\hline
MagMonitor \cite{feng2020magmonitor}& N/A / N/A & \makecell*[l]{Image, and\\ Histogram of Orie-\\nted Gradients\\ (HOG)} & \blue{RBF-Kernel} SVM & \makecell*[l]{Sedan (93.1/95.0),\\ SUV and Van (85.0/79.0),\\ Bus (93.2/96.0), and\\ Truck (89.2/91.0)} &  87.0  \\
\hline
\textbf{\TheName } &{1000 / 1153}  & \makecell*[l]{9 Features listed\\ in Table~\ref{table:Features}} & \blue{RBF-Kernel} SVM &\makecell*[l]{Motorbike~(100.0/100.0),\\ Sedan and SUV~(97.7/93.8),\\ Light Truck~(92.3/92.3),\\ Bus (70.2/86.8),\\ Medium Truck (88.7/77.6),\\ Heavy Truck (46.4/72.2), and\\ Super Truck (92.0/95.8)}& {90.3} \\
\hline
\end{tabular}
\end{table*}
\subsection{Overall Results and Comparisons}
With data collected, we carried out the experiments using the proposed algorithms and compare results in three folders.

\emph{Vehicle Length Classification - }As was discussed, \TheName\ classifies the type of vehicles according to their lengths and other signal features. In this part, we first intend to understand the accuracy of length classification.  Table~\ref{table:4types} presents the confusion matrix for vehicle length classification. It is obvious that \TheName\ achieves high accuracy in vehicle length classification, as only 54 samples in all 1153 test cases were misclassified ($\approx 4.7\%$). The major difficulty of vehicle length classification lays on the discriminant between the vehicles in the range of $(3,6]m$ and $(6,12]m$. As was mentioned, a semi-automated learning paradigm has been used to search the best parameters for highpass/lowpass filtering and the length estimator. It should be noted that the discriminability of vehicle types between ``Sedan and SUV'', ``Light Truck'', ``Bus'', ``Medium Truck'', and ``Heavy Truck'' using \TheName\ would be bounded by the vehicle length classification.

\emph{Vehicle Type Classification - } With vehicle length classified, \TheName\ enables the vehicle type classification using a hierarchical recognition model. Table~\ref{table:7types} presents the confusion matrix of vehicle type classification using \TheName. It is obvious that \TheName\ achieves decent accuracy in vehicle type classification, around 10\% testing samples have been misclassified. The  misclassification majorly happens between ``Sedan and SUV'', ``Light Truck'', ``Bus'', ``Medium Truck'', and ``Heavy Truck'', partially due to the indiscriminability of signals in vehicle length estimation. In addition to the classification error caused by length estimation, it is often difficult to classify the types of vehicles in the same range of length. For example, 20 of 806 testing samples were misclassified between ``Sedan and SUV'' and ``Light Truck'' (both of them are in $(3,6]m$ length), while, in the categories of ``Bus'', ``Medium Truck'', and ``Heavy Truck'', 38 testing samples were misclassified. In this way, we could conclude that due to the misclassification of vehicle lengths, there were 54 out of 1153 vehicles in the testing enviornment were misclassified (the first layer of the hierarchical model in Figure~\ref{fig:decisionTree}); due to the misclassification based on the frequency and temporal features of extracted signals, there were 58 vehicles were misclassified (the second layer of the hierarchical model in Figure~\ref{fig:decisionTree}). In an overall manner, there are totally 112 testing samples were misclassified and the overall accuracy is 90.3\%.

\emph{Comparisons to Existing Systems - } To the proposed system, we compare \TheName\ with some recent magnetic-based sensing systems for vehicle monitoring, including Xu et al.~\cite{xu2017vehicle} (published in 2017), SMOTE~\cite{xu2018vehicle} (published in 2018), and MagMonitor~\cite{feng2020magmonitor} (published in 2020). Table~\ref{table:Solution_Comparison} presents the details in comparisons, including number of training and testing samples, features and classifiers used, accuracy details per vehicle type, and the overall classification accuracy. To understand the classification accuracy for every type of vehicle, we calculate the precision and recall for classifying every type of vehicle.  Though the overall accuracy of \TheName\ is slightly lower than the accuracy reported by SMOTE~\cite{xu2018vehicle}, the proposed system can classify vehicles into more fine-grained types. In this way, while these solutions sort vehicles into different categories, we could conclude that (1) \TheName\ enables vehicle type classification in an even finer granularity (7 types) compared to other solutions, and (2) \TheName\ achieves comparable classification accuracy in all the seven types.



\subsection{Ablation Studies and Analysis}
To further understand \TheName, we carried out extensive ablation studies to evaluate effectiveness of features and classifiers for the vehicle type classification. 

\emph{Using Temporal Domain Features Only - } In this ablation study, we evaluate the accuracy of vehicle type classification using support vector machine with Temporal Domain Features only. In general, we follow the hierarchical recognition model shown in Figure~\ref{fig:decisionTree} to obtain the vehicle length classification results. With the vehicle length predicted, we use the temporal domain features listed in Table~\ref{table:Features} to classify the vehicle types. In this way, the classification results for ``Motorbike'' and ``Super Truck'' would not be changed. The confusion matrix of vehicle type classification using Temporal Domain Features is listed in Table~\ref{table:7types-temp}. It is obvious that the solution based on Temporal Domain Features only suffers the serious performance degradation. Specifically, the algorithm in this setting proportionally misclassifies: ``Light Truck'' into the type of ``Sedan and SUV'', ``Medium Truck'' into the type of ``Bus'', and ``Bus'' into the type of ``Medium Truck''. 

\begin{table*}
\caption{Confusion Matrix of Vehicle Type Classification using Temporal Domain Features.}
\label{table:7types-temp}
\centering
\begin{tabular}{|c|c|c|c|c|c|c|c|}
\hline
\diagbox{Truth}{Predicted}  & Motorbike & Sedan and SUV & Light Truck & Bus & Medium Truck & Heavy Truck & Super Truck \\
\hline
Motorbike & \textbf{15}  & \red{0} & \red{0}  & \red{0}  & \red{0} & \red{0} & \red{0} \\
\hline
Sedan and SUV & \red{0} & \textbf{506}  & \red{5}  & \red{5}  & \red{7} & \red{9} & \red{0} \\
\hline
Light Truck & \red{0} & \red{233} & \textbf{62}  & \red{5} & \red{0} & \red{9} & \red{0} \\
\hline
Bus & \red{0} & \red{4} & \red{0} & \textbf{8}  & \red{60} & \red{4} & \red{0} \\
\hline
Medium Truck & \red{0} & \red{2} & \red{5} & \red{25} & \textbf{31} & \red{97} & \red{1} \\
\hline
Heavy Truck & \red{0} & \red{1} & \red{4} & \red{1} & \red{1} & \textbf{28} & \red{1} \\
\hline
Super Truck & \red{0} & \red{0} & \red{0} & \red{0} & \red{0} & \red{1} & \textbf{23} \\
\hline
\end{tabular}
\end{table*}

\emph{Using Frequency Domain Features Only - } In this ablation study, we evaluate the accuracy of vehicle type classification using support vector machine with Frequency Domain Features only. We also follow the hierarchical recognition model shown in Figure~\ref{fig:decisionTree} and replace the three classifiers in the second layer with ones based on the Frequency Domain Features (listed in Table~\ref{table:Features}). The confusion matrix of vehicle type classification using Temporal Domain Features is listed in Table~\ref{table:7types-freq}. It is obvious that the solution based on Frequency Domain Features only suffers the serious performance degradation. Specifically, the algorithm in this setting proportionally misclassifies: ``Sedan and SUV'' into the types of ``Light Truck'' and ``Bus'', ``Light Truck'' into the type of ``Sedan and SUV'', ``Bus'' into the type of the type of ``Medium Truck'', and ``Medium Truck'' into the type of ``Heavy Truck''. 

\begin{table*}
\caption{Confusion Matrix of Vehicle Type Classification using Frequency Domain Features.}
\label{table:7types-freq}
\centering
\begin{tabular}{|c|c|c|c|c|c|c|c|}
\hline
\diagbox{Truth}{Predicted}  & Motorbike & Sedan and SUV & Light Truck & Bus & Medium Truck & Heavy Truck & Super Truck \\
\hline
Motorbike & \textbf{15}  & \red{0} & \red{0}  & \red{0}  & \red{0} & \red{0} & \red{0} \\
\hline
Sedan and SUV & \red{0} & \textbf{506}  & \red{5}  & \red{15}  & \red{2} & \red{4} & \red{0} \\
\hline
Light Truck & \red{0} & \red{224} & \textbf{71}  & \red{9} & \red{1} & \red{4} & \red{0} \\
\hline
Bus & \red{0} & \red{2} & \red{2} & \textbf{29}  & \red{3} & \red{13} & \red{0} \\
\hline
Medium Truck & \red{0} & \red{1} & \red{6} & \red{90} & \textbf{61} & \red{2} & \red{1} \\
\hline
Heavy Truck & \red{0} & \red{1} & \red{4} & \red{2} & \red{1} & \textbf{27} & \red{1} \\
\hline
Super Truck & \red{0} & \red{0} & \red{0} & \red{0} & \red{0} & \red{1} & \textbf{23} \\
\hline
\end{tabular}
\end{table*}

\emph{Using Other Classifiers - } In our work, we select  \blue{RBF-Kernel} SVM as the classifiers through a simple automated learning procedure via the comparisons of cross-validation accuracy. To understand the superiority of \blue{RBF-Kernel} SVM in \TheName\ tasks, we replace \blue{RBF-Kernel} SVM learners used in \TheName\ with other classifiers, including Random Forest, Adaboost, Neural Network, and so on. In all settings, we find significant performance degradation in terms of overall classification accuracy. The second best learner for our task is the Random Forest Classifier. Table~\ref{table:7types-rand} demonstrates the confusion matrix of vehicle type classification using Random Forest Classifiers with both Temporal and Frequency Domain Features. Compared to \TheName, Random Forest Classifier might cause more errors in classifying vehicular types between ``Sedan and SUV'' and ``Light Truck'', ``Bus'' and ``Medium Truck'', ``Medium Truck'' and ``Heavy Truck'' obviously.

\blue{We believe RBF-Kernel SVM outperforms other algorithms due to two reasons as follows. (I) RBF-Kernel SVM leverages kernel tricks to project the data in orginal feature space into a high-dimensional nonlinear kernel space, where the samples are more discriminative and distinguishable; and (II) SVM classifier adopts a max-margin loss that aims at minimizing the confusion between the overlapped types (e.g., ``Medium Truck'' and ``Heavy Truck'')~\cite{scholkopf2005learning}.}

\begin{table*}
\caption{Confusion Matrix of Vehicle Type Classification using both Time and Frequency Domain Features and Random Forest Classifiers.}
\label{table:7types-rand}
\centering
\begin{tabular}{|c|c|c|c|c|c|c|c|}
\hline
\diagbox{Truth}{Predicted} & Motorbike & Sedan and SUV & Light Truck & Bus & Medium Truck & Heavy Truck & Super Truck \\
\hline
Motorbike & \textbf{15}  & \red{0} & \red{0}  & \red{0}  & \red{0} & \red{0} & \red{0} \\
\hline
Sedan and SUV & \red{0} & \textbf{462}  & \red{49}  & \red{10}  & \red{9} & \red{2} & \red{0} \\
\hline
Light Truck & \red{0} & \red{75} & \textbf{220}  & \red{8} & \red{0} & \red{6} & \red{0} \\
\hline
Bus & \red{0} & \red{1} & \red{3} & \textbf{43}  & \red{27} & \red{2} & \red{0} \\
\hline
Medium Truck & \red{0} & \red{7} & \red{0} & \red{1} & \textbf{92} & \red{60} & \red{1} \\
\hline
Heavy Truck & \red{0} & \red{3} & \red{2} & \red{1} & \red{2} & \textbf{27} & \red{1} \\
\hline
Super Truck & \red{0} & \red{0} & \red{0} & \red{0} & \red{0} & \red{1} & \textbf{23} \\
\hline
\end{tabular}
\end{table*}

\blue{\subsection{Time Consumption Analysis}}
\blue{We analyze the runtime records of all 1153 vehicles in the experiment and estimate the end-to-end time consumption (from detecting a passing vehicle to classifying its type) for every vehicle. For all type of vehicles, the average time consumption of \TheName\ for recognize one vehicle is 0.0995$\pm$0.0183 seconds. We believe the average measure of time consumption is reliable, as the Variance-to-Mean Ratio (VMR) is $0.018^2/0.0995=0.32\%$ close to zero (i.e., small dispersion against a stable mean).  }

\blue{To provide an estimate of Worst Case Execution Time (WCET)~\cite{abella2017measurement}, we calculate the confidence upper bound of the time consumption using the Six-Sigma standard for extreme value estimates~\cite{aleksandrovskaya2019application}, which should be $0.0995+6.0\times 0.0183=0.2093$ seconds in our experiments. Note that in the normal traffic condition of a highway, vehicles on every lane should pass by the sensor nodes sequentially in a one-by-one manner. Moreover, in the practice, it frequently recommends a minimum time gap of 2.0 between two consecutive vehicles for driving safety~\cite{yimer2020study}. In this way, we could conclude that the time consumption of \TheName\ indeed ensures the real-time performance of traffic monitoring. }





\section{Discussion and Limitations}
There are several limitations in our study. Here We discuss some of the limitations and technical issues as follow.
    
\paragraph{Hierarchical Recognition Models}~Instead of proposing an end-to-end model for classifying the vehicle types from raw signals, \TheName\ recognizes the vehicle types in three steps: (1) speed/velocity estimation, (2) vehicle length estimation/classification (in 4 categories), and (3) vehicle type classification (in 7 categories). The proposed algorithm push the granularity of vehicle type classification finer and finer. In our experiments, totally 112 out of 1153 (9.7\%) testing samples were misclassified. While 48.2\%  (54 out of 112) errors were due to the misclassification of vehicle length, the rest 51.8\% was caused by the vehicle type classification using Temporal/Frequency Domain Features. It is reasonable to assume the use of some end-to-end approach based on raw signal features could outperform the proposed solutions. However, compared to existing work~\cite{xu2017vehicle,xu2018vehicle,feng2020magmonitor}, the overall accuracy of \TheName\ tops while the types/categorization of vehicles are finest.

\blue{Note that, we consider the estimated speeds and lengths of vehicles as two key features for the hierarchical recognition framework due to two reasons as follows. (I) The types of vehicles are majorly categorized by the lengths of vehicles according to the standard recommended by transportation administration and authorities~\cite{TPRI}; and (II) it frequently needs to first measure the speed of a vehicle and further estimate its length using the measured speed and the measured time duration for the vehicle to drive through the two sensor nodes. Of-course, other features including heights and weights of vehicles might also help for classification. We consider the proposed \TheName\ as an alternative solution of the problem, which could complement with other existing tools and systems for better overall performance. }

\paragraph{Validation of Speed/Velocity Estimation}~The vehicle length and type classification of \TheName\ majorly relies on the estimation of vehicle speed/velocity. In our research, we collect vehicle data under real-world deployment of \TheName\ system using magnetic sensors on the road surface and video recorders survelling the road section. We recruit professional labellers to recognize the models of vehicles, and label the collected signals with the vehicle types and lengths. In this way, we did not collect the real-time speed/velocity of vehicles as the label of data, and thus cannot evaluate the accuracy of speed/velocity estimation. All in all, \TheName\ infers the length of vehicle using the estimated speed through a white-box physical model (in Eq.~\ref{equ:vehiclelength}). As the overall vehicle length/type classification is accurate, we can conclude that the speed/velocity estimation of \TheName\ did not cause significant performance degradation for the further steps.

\paragraph{Shallow Models and Deep Learning}~\TheName\ employs an semi-automated procedure to extract Temporal/Frequency Domain Features from raw magnetic signals, and adopts statistical models/learners are used to handle the classification tasks. Because the features used in \TheName\ are with relatively low dimensions.  It is no doubt that the use of some deep neural networks could achieve better performance when stacking feature learning and discriminative learning in an end-to-end optimization manner. The contribution of our study is to demonstrate the feasibility of using a hierarchical recognition model that infer the type of vehicles step-by-step from speed estimation, to length estimation, and finally to the vehicle type classification, in an interpretable way.

\paragraph{Sensors and Vehicles Coordination}~To estimate speed, \TheName\ assumes that the vehicles would go straight through the two magnetic sensors, and the speed/velocity of the vehicle should be in a certain range with respect to the sampling rate of ADC and the distance between sensors. In fact, \TheName\ deploys sensors on the straight section of highways and the distance between two sensor nodes is $d=1m$ in our experiments. In this way, we are pretty sure that the vehicles should pass the two sensors in a straight line. As the sampling frequency is set to $f=1000$~Hz , \TheName\ can well detect the vehicles going through the sensor nodes with a velocity in the range of $(20,~150]$~km/h. \blue{Moreover, the performance of \TheName\ could be improved using the Vehicle-to-Vehicle (VoV) and Vehicle-to-Infrastructure (VoI) communications~\cite{kong2017millimeter}.}

\blue{\paragraph{Weathers, Noises, and Other Factors may affect the performance} Some environmental factors, including extreme weather conditions, electromagnetic interference, vibration of road surfaces, and alternating current (AC) harmonics, would affect the working conditions of magnetic sensors and \TheName. For the noises caused by harmonics in AC power, vibration of road surfaces, and electromagnetic interference, \TheName\ tries to eliminate the effects of noises to classification results through filtering. Specifically, \TheName\ adopts a well-designed bandstop filter to remove the effects of harmonics and the power frequency (e.g., 50 Hz in China and Europe) in AC. Then, a pair of highpass and lowpass filters has used to remove the effects of electromagnetic interference in some bands. Further, \TheName\ considers the vibration of road surfaces (especially when heavy vehicles are approaching or departing) and tries to remove the effects of road surface vibration through filtering the temporal signals in the ``fade-in'' and ``fade-out'' periods. Note that the parameters for highpass/lowpass filters and fade-in/fade-out periods detection are all fine-tuned with automated learning techniques with appropriate validations. Finally, to handle extreme weather conditions, it is highly recommended to complement \TheName\ with other traffic monitoring tools (e.g., vision-based or radar-based solutions) to achieve good performance in general.}

\section{Conclusion}
The operation and management of intelligent transportation systems (ITS) relies on the real-time recognition of vehicle types (e.g., cars, trucks, and buses), in the critical roads and highways. In our research, we propose \TheName, that recognizes the types of running vehicles using a pair of magnetic sensor nodes deployed on the surface of road section. Specifically, \TheName\ filters out noises, and segments received magnetic signals by the time points that the vehicle arrives or departs from every sensor node. Further, \TheName\ adopts a hierarchical recognition model to first estimate the speed/velocity, and latter identify the length of vehicle using the predicted speed and the distance between the sensor nodes. With the vehicle length identified and the temporal/spectral features extracted from the magnetic signals, \TheName\ classify the types of vehicles accordingly. More specifically, \TheName\ adopts semi-automated learning techniques to optimize the parameters of lowpass/highpass filters, design of features, and the choice of hyper-parameters for length estimation. Extensive experiment based on real-word field deployment (on the highways in Shenzhen, China) shows that \TheName\ significantly outperforms the existing methods in both classification accuracy and the granularity of vehicle types (i.e., 7 types by \TheName\ versus 4 types by the existing work~\cite{xu2017vehicle,xu2018vehicle,feng2020magmonitor}). To be specific, our field experiment results validate that \TheName\ is with at least $90\%$ vehicle type classification accuracy and less than 5\% vehicle length classification error.

\section*{Acknowledgement}
This work is supported in part by National Key R\&D Program of China (No. 2019YFB2102100 to Kafeng Wang and Cheng-Zhong Xu/No. 2018YFB1402600 to Haoyi Xiong). The collaboration between University of Macau and Baidu Research is sponsored by Science and Technology Development Fund of Macao S.A.R (FDCT) under number 0015/2019/AKP. This work is also supported in part by Shenzhen Engineering Research Center for Beidou Positioning Service Technology No.XMHT20190101035. H. Chen is supported in part by the funding project of Zhejiang Lab under Grant 2020LC0PI01.





%

\bibliography{vehicleRef}

\begin{thebibliography}{10}
\providecommand{\url}[1]{#1}
\csname url@samestyle\endcsname
\providecommand{\newblock}{\relax}
\providecommand{\bibinfo}[2]{#2}
\providecommand{\BIBentrySTDinterwordspacing}{\spaceskip=0pt\relax}
\providecommand{\BIBentryALTinterwordstretchfactor}{4}
\providecommand{\BIBentryALTinterwordspacing}{\spaceskip=\fontdimen2\font plus
\BIBentryALTinterwordstretchfactor\fontdimen3\font minus
  \fontdimen4\font\relax}
\providecommand{\BIBforeignlanguage}[2]{{%
\expandafter\ifx\csname l@#1\endcsname\relax
\typeout{** WARNING: IEEEtran.bst: No hyphenation pattern has been}%
\typeout{** loaded for the language `#1'. Using the pattern for}%
\typeout{** the default language instead.}%
\else
\language=\csname l@#1\endcsname
\fi
#2}}
\providecommand{\BIBdecl}{\relax}
\BIBdecl

\bibitem{zhang2011data}
J.~Zhang, F.-Y. Wang, K.~Wang, W.-H. Lin, X.~Xu, and C.~Chen, ``Data-driven
  intelligent transportation systems: A survey,'' \emph{IEEE Transactions on
  Intelligent Transportation Systems}, vol.~12, no.~4, pp. 1624--1639, 2011.

\bibitem{tian2014hierarchical}
B.~Tian, B.~T. Morris, M.~Tang, Y.~Liu, Y.~Yao, C.~Gou, D.~Shen, and S.~Tang,
  ``Hierarchical and networked vehicle surveillance in its: a survey,''
  \emph{IEEE transactions on intelligent transportation systems}, vol.~16,
  no.~2, pp. 557--580, 2014.

\bibitem{miller2008vehicle}
J.~Miller, ``Vehicle-to-vehicle-to-infrastructure (v2v2i) intelligent
  transportation system architecture,'' in \emph{2008 IEEE intelligent vehicles
  symposium}.\hskip 1em plus 0.5em minus 0.4em\relax IEEE, 2008, pp. 715--720.

\bibitem{milanes2012intelligent}
V.~Milanes, J.~Villagra, J.~Godoy, J.~Simo, J.~P{\'e}rez, and E.~Onieva, ``An
  intelligent v2i-based traffic management system,'' \emph{IEEE Transactions on
  Intelligent Transportation Systems}, vol.~13, no.~1, pp. 49--58, 2012.

\bibitem{el2018towards}
A.~S. El-Wakeel, J.~Li, A.~Noureldin, H.~S. Hassanein, and N.~Zorba, ``Towards
  a practical crowdsensing system for road surface conditions monitoring,''
  \emph{IEEE Internet of Things Journal}, vol.~5, no.~6, pp. 4672--4685, 2018.

\bibitem{silva2013traffic}
T.~H. Silva, P.~O.~V. De~Melo, A.~C. Viana, J.~M. Almeida, J.~Salles, and A.~A.
  Loureiro, ``Traffic condition is more than colored lines on a map:
  characterization of waze alerts,'' in \emph{International Conference on
  Social Informatics}.\hskip 1em plus 0.5em minus 0.4em\relax Springer, 2013,
  pp. 309--318.

\bibitem{zhang20144w1h}
D.~Zhang, L.~Wang, H.~Xiong, and B.~Guo, ``4w1h in mobile crowd sensing,''
  \emph{IEEE Communications Magazine}, vol.~52, no.~8, pp. 42--48, 2014.

\bibitem{xiong2016sensus}
H.~Xiong, Y.~Huang, L.~E. Barnes, and M.~S. Gerber, ``Sensus: a cross-platform,
  general-purpose system for mobile crowdsensing in human-subject studies,'' in
  \emph{Proceedings of the 2016 ACM international joint conference on pervasive
  and ubiquitous computing}.\hskip 1em plus 0.5em minus 0.4em\relax ACM, 2016,
  pp. 415--426.

\bibitem{xiong2015crowdtasker}
H.~Xiong, D.~Zhang, G.~Chen, L.~Wang, and V.~Gauthier, ``Crowdtasker:
  Maximizing coverage quality in piggyback crowdsensing under budget
  constraint,'' in \emph{2015 IEEE International Conference on Pervasive
  Computing and Communications (PerCom)}.\hskip 1em plus 0.5em minus
  0.4em\relax IEEE, 2015, pp. 55--62.

\bibitem{hofstede2009surfmap}
R.~Hofstede and T.~Fioreze, ``Surfmap: A network monitoring tool based on the
  google maps api,'' in \emph{2009 IFIP/IEEE International Symposium on
  Integrated Network Management}.\hskip 1em plus 0.5em minus 0.4em\relax IEEE,
  2009, pp. 676--690.

\bibitem{burney2017google}
A.~Burney, M.~Asif, Z.~Abbas, and S.~Burney, ``Google maps security concerns,''
  \emph{Journal of Computer and Communications}, vol.~6, no.~1, pp. 275--283,
  2017.

\bibitem{Caruso1999Vehicle}
M.~J. Caruso and L.~S. Withanawasam, ``Vehicle detection and compass
  applications using amr magnetic sensors,'' in \emph{Sensors Expo
  Proceedings}, vol. 477, 1999, p.~39.

\bibitem{Cheung2007Traffic}
S.~Y. Cheung and P.~Varaiya, ``Traffic surveillance by wireless sensor
  networks: Final report,'' \emph{Path Research Report}, 2007.

\bibitem{zhang2010distributed}
W.~Zhang, G.~Tan, H.-M. Shi, and M.-W. Lin, ``A distributed threshold algorithm
  for vehicle classification based on binary proximity sensors and intelligent
  neuron classifier.'' \emph{J. Inf. Sci. Eng.}, vol.~26, no.~3, pp. 769--783,
  2010.

\bibitem{zhang2011real}
L.~Zhang, R.~Wang, L.~Cui \emph{et~al.}, ``Real-time traffic monitoring with
  magnetic sensor networks,'' \emph{Journal of information science and
  engineering}, vol.~27, no.~4, pp. 1473--1486, 2011.

\bibitem{Xu2016Traffic}
B.~Xu, H.~Li, D.~Xu, L.~Jia, B.~Ran, and J.~Rong, ``Traffic vehicle counting in
  jam flow conditions using low-cost and energy-efficient wireless magnetic
  sensors:,'' \emph{Sensors}, vol.~16, no.~11, p. 1868, 2016.

\bibitem{xu2017vehicle}
C.~Xu, Y.~Wang, and Y.~Zhan, ``Vehicle classification under different feature
  sets with a single anisotropic magnetoresistive sensor,'' \emph{IEICE
  TRANSACTIONS on Fundamentals of Electronics, Communications and Computer
  Sciences}, vol. 100, no.~2, pp. 440--447, 2017.

\bibitem{xu2018vehicle}
C.~Xu, Y.~Wang, X.~Bao, and F.~Li, ``Vehicle classification using an imbalanced
  dataset based on a single magnetic sensor,'' \emph{Sensors}, vol.~18, no.~6,
  p. 1690, 2018.

\bibitem{feng2020magmonitor}
Y.~Feng, G.~Mao, B.~Cheng, C.~Li, Y.~Hui, Z.~Xu, and J.~Chen, ``Magmonitor:
  Vehicle speed estimation and vehicle classification through a magnetic
  sensor,'' \emph{IEEE Transactions on Intelligent Transportation Systems},
  2020.

\bibitem{minge2012loop}
E.~D. Minge, S.~Peterson, H.~Weinblatt, B.~Coifman, and E.~Hoekman, ``Loop-and
  length-based vehicle classification: Federal highway administration-pooled
  fund program [tpf-5 (192)],'' Minnesota Department of Transportation,
  Research Services, Tech. Rep., 2012.

\bibitem{TPRI}
T.~Planning and C.~Research~Institute, Ministry of~Transport, ``Tpri,''
  \url{http://http://www.tpri.org.cn}.

\bibitem{liu2019vehicle}
K.~Liu, Z.~Asher, X.~Gong, M.~Huang, and I.~Kolmanovsky, ``Vehicle velocity
  prediction and energy management strategy part 1: Deterministic and
  stochastic vehicle velocity prediction using machine learning,'' SAE
  Technical Paper, Tech. Rep., 2019.

\bibitem{wahlstrom2014magnetometer}
N.~Wahlstr{\"o}m and F.~Gustafsson, ``Magnetometer modeling and validation for
  tracking metallic targets,'' \emph{IEEE Transactions on Signal Processing},
  vol.~62, no.~3, pp. 545--556, 2014.

\bibitem{lan2009vehicle}
J.~Lan and Y.~Shi, ``Vehicle detection and recognition based on a mems magnetic
  sensor,'' in \emph{2009 4th IEEE International Conference on Nano/Micro
  Engineered and Molecular Systems}.\hskip 1em plus 0.5em minus 0.4em\relax
  IEEE, 2009, pp. 404--408.

\bibitem{jolevski2011smart}
I.~Jolevski, A.~Markoski, and R.~Pasic, ``Smart vehicle sensing and
  classification node with energy aware vehicle classification algorithm,'' in
  \emph{Proceedings of the ITI 2011, 33rd International Conference on
  Information Technology Interfaces}.\hskip 1em plus 0.5em minus 0.4em\relax
  IEEE, 2011, pp. 409--414.

\bibitem{balid2017intelligent}
W.~Balid, H.~Tafish, and H.~H. Refai, ``Intelligent vehicle counting and
  classification sensor for real-time traffic surveillance,'' \emph{IEEE
  Transactions on Intelligent Transportation Systems}, vol.~19, no.~6, pp.
  1784--1794, 2017.

\bibitem{balid2015development}
------, ``Development of portable wireless sensor network system for real-time
  traffic surveillance,'' in \emph{2015 IEEE 18th International Conference on
  Intelligent Transportation Systems}.\hskip 1em plus 0.5em minus 0.4em\relax
  IEEE, 2015, pp. 1630--1637.

\bibitem{gontarz2015use}
S.~Gontarz, P.~Szulim, J.~Se{\'n}ko, and J.~Dyba{\l}a, ``Use of magnetic
  monitoring of vehicles for proactive strategy development,''
  \emph{Transportation Research Part C: Emerging Technologies}, vol.~52, pp.
  102--115, 2015.

\bibitem{bitar2016probabilistic}
N.~Bitar and H.~H. Refai, ``A probabilistic approach to improve the accuracy of
  axle-based automatic vehicle classifiers,'' \emph{IEEE Transactions on
  Intelligent Transportation Systems}, vol.~18, no.~3, pp. 537--544, 2016.

\bibitem{wang2017roadside}
Q.~Wang, J.~Zheng, H.~Xu, B.~Xu, and R.~Chen, ``Roadside magnetic sensor system
  for vehicle detection in urban environments,'' \emph{IEEE Transactions on
  Intelligent Transportation Systems}, vol.~19, no.~5, pp. 1365--1374, 2017.

\bibitem{kleyko2015comparison}
D.~Kleyko, R.~Hostettler, W.~Birk, and E.~Osipov, ``Comparison of machine
  learning techniques for vehicle classification using road side sensors,'' in
  \emph{2015 IEEE 18th International Conference on Intelligent Transportation
  Systems}.\hskip 1em plus 0.5em minus 0.4em\relax IEEE, 2015, pp. 572--577.

\bibitem{xie2014simulations}
J.~Xie, C.~Qin, X.~Zhou, L.~Huang, X.~Han, M.~Wang, and L.~Li, ``The
  simulations and experiments of the electromagnetic tracking system based on
  magnetic dipole model,'' \emph{IEEE transactions on applied
  superconductivity}, vol.~24, no.~3, pp. 1--4, 2014.

\bibitem{zhou2012practicable}
Q.~Zhou, G.~Tong, B.~Li, and X.~Yuan, ``A practicable method for ferromagnetic
  object moving direction identification,'' \emph{IEEE Transactions on
  Magnetics}, vol.~48, no.~8, pp. 2340--2345, 2012.

\bibitem{gontarz2011impact}
S.~Gontarz and S.~Radkowski, ``Impact of various factors on relationships
  between stress and eigen magnetic field in a steel specimen,'' \emph{IEEE
  Transactions on magnetics}, vol.~48, no.~3, pp. 1143--1154, 2011.

\bibitem{ren2015magnetic}
Y.~Ren, C.~Hu, S.~Xiang, and Z.~Feng, ``Magnetic dipole model in the
  near-field,'' in \emph{2015 IEEE International Conference on Information and
  Automation}.\hskip 1em plus 0.5em minus 0.4em\relax IEEE, 2015, pp.
  1085--1089.

\bibitem{taghvaeeyan2013portable}
S.~Taghvaeeyan and R.~Rajamani, ``Portable roadside sensors for vehicle
  counting, classification, and speed measurement,'' \emph{IEEE Transactions on
  Intelligent Transportation Systems}, vol.~15, no.~1, pp. 73--83, 2014.

\bibitem{Zhang2015A}
Z.~Zhang, T.~Zhao, and H.~Yuan, \emph{A Vehicle Speed Estimation Algorithm
  Based on Wireless AMR Sensors}.\hskip 1em plus 0.5em minus 0.4em\relax
  Springer International Publishing, 2015.

\bibitem{wei2017adaptable}
Q.~Wei and B.~Yang, ``Adaptable vehicle detection and speed estimation for
  changeable urban traffic with anisotropic magnetoresistive sensors,''
  \emph{IEEE Sensors Journal}, vol.~17, no.~7, pp. 2021--2028, 2017.

\bibitem{haoui2008wireless}
A.~Haoui, R.~Kavaler, and P.~Varaiya, ``Wireless magnetic sensors for traffic
  surveillance,'' \emph{Transportation Research Part C: Emerging Technologies},
  vol.~16, no.~3, pp. 294--306, 2008.

\bibitem{obertov2014vehicle}
D.~Obertov, V.~Bardov, and B.~Andrievsky, ``Vehicle speed estimation using
  roadside sensors,'' in \emph{2014 6th International Congress on Ultra Modern
  Telecommunications and Control Systems and Workshops (ICUMT)}.\hskip 1em plus
  0.5em minus 0.4em\relax IEEE, 2014, pp. 111--117.

\bibitem{li2011some}
H.~Li, H.~Dong, L.~Jia, D.~Xu, and Y.~Qin, ``Some practical vehicle speed
  estimation methods by a single traffic magnetic sensor,'' in \emph{2011 14th
  International IEEE Conference on Intelligent Transportation Systems
  (ITSC)}.\hskip 1em plus 0.5em minus 0.4em\relax IEEE, 2011, pp. 1566--1573.

\bibitem{xiaoyong2010vehicle}
X.~DENG, Z.~HU, Z.~Peng, and J.~GUO, ``Vehicle class composition identification
  based mean speed estimation algorithm using single magnetic sensor,''
  \emph{Journal of Transportation Systems Engineering and Information
  Technology}, vol.~10, no.~5, pp. 35--39, 2010.

\bibitem{feng2019magspeed}
Y.~Feng, G.~Mao, B.~Cheng, B.~Huang, S.~Wang, and J.~Chen, ``Magspeed: A novel
  method of vehicle speed estimation through a single magnetic sensor,'' in
  \emph{2019 IEEE Intelligent Transportation Systems Conference (ITSC)}.\hskip
  1em plus 0.5em minus 0.4em\relax IEEE, 2019, pp. 4281--4286.

\bibitem{chen2019roadside}
Z.~Chen, Z.~Liu, Y.~Hui, W.~Li, C.~Li, T.~H. Luan, and G.~Mao, ``Roadside
  sensor based vehicle counting incomplex traffic environment,'' in \emph{2019
  IEEE Globecom Workshops (GC Wkshps)}.\hskip 1em plus 0.5em minus 0.4em\relax
  IEEE, 2019, pp. 1--5.

\bibitem{Kaewkamnerd2010Vehicle}
S.~Kaewkamnerd, J.~Chinrungrueng, R.~Pongthornseri, and S.~Dumnin, ``Vehicle
  classification based on magnetic sensor signal,'' in \emph{IEEE International
  Conference on Information and Automation}, 2010, pp. 935--939.

\bibitem{de2016simple}
G.~De~Angelis, A.~De~Angelis, V.~Pasku, A.~Moschitta, and P.~Carbone, ``A
  simple magnetic signature vehicles detection and classification system for
  smart cities,'' in \emph{2016 IEEE International Symposium on Systems
  Engineering (ISSE)}.\hskip 1em plus 0.5em minus 0.4em\relax IEEE, 2016, pp.
  1--6.

\bibitem{kong2017millimeter}
L.~{Kong}, M.~K. {Khan}, F.~{Wu}, G.~{Chen}, and P.~{Zeng}, ``Millimeter-wave
  wireless communications for iot-cloud supported autonomous vehicles:
  Overview, design, and challenges,'' \emph{IEEE Communications Magazine},
  vol.~55, no.~1, pp. 62--68, 2017.

\bibitem{velisavljevic2016wireless}
V.~Velisavljevic, E.~Cano, V.~Dyo, and B.~Allen, ``Wireless magnetic sensor
  network for road traffic monitoring and vehicle classification,''
  \emph{Transport and Telecommunication Journal}, vol.~17, no.~4, pp. 274--288,
  2016.

\bibitem{nargesian2017learning}
F.~Nargesian, H.~Samulowitz, U.~Khurana, E.~B. Khalil, and D.~S. Turaga,
  ``Learning feature engineering for classification.'' in \emph{IJCAI}, 2017,
  pp. 2529--2535.

\bibitem{chen2019neural}
X.~Chen, Q.~Lin, C.~Luo, X.~Li, H.~Zhang, Y.~Xu, Y.~Dang, K.~Sui, X.~Zhang,
  B.~Qiao \emph{et~al.}, ``Neural feature search: A neural architecture for
  automated feature engineering,'' in \emph{2019 IEEE International Conference
  on Data Mining (ICDM)}.\hskip 1em plus 0.5em minus 0.4em\relax IEEE, 2019,
  pp. 71--80.

\bibitem{kaul2017autolearn}
A.~Kaul, S.~Maheshwary, and V.~Pudi, ``Autolearn—automated feature generation
  and selection,'' in \emph{2017 IEEE International Conference on data mining
  (ICDM)}.\hskip 1em plus 0.5em minus 0.4em\relax IEEE, 2017, pp. 217--226.

\bibitem{kotthoff2017auto}
L.~Kotthoff, C.~Thornton, H.~H. Hoos, F.~Hutter, and K.~Leyton-Brown,
  ``Auto-weka 2.0: Automatic model selection and hyperparameter optimization in
  weka,'' \emph{The Journal of Machine Learning Research}, vol.~18, no.~1, pp.
  826--830, 2017.

\bibitem{chapelle1999support}
O.~{Chapelle}, P.~{Haffner}, and V.~{Vapnik}, ``Support vector machines for
  histogram-based image classification,'' \emph{IEEE Transactions on Neural
  Networks}, vol.~10, no.~5, pp. 1055--1064, 1999.

\bibitem{scholkopf2005learning}
B.~Scholkopf and J.~A. Smola, ``Learning with kernels: Support vector machines,
  regularization, optimization, and beyond,'' \emph{IEEE Transactions on Neural
  Networks}, pp. 781--781, 2005.

\bibitem{abella2017measurement}
J.~Abella, M.~Padilla, D.~J. Castillo, and J.~F. Cazorla, ``Measurement-based
  worst-case execution time estimation using the coefficient of variation,''
  \emph{ACM Trans. Design Autom. Electr. Syst.}, pp. 72:1--72:29, 2017.

\bibitem{aleksandrovskaya2019application}
Lidiya N. Aleksandrovskaya, Anna E. Ardalionova, and Ljubisa Papic,
  ``Application of probability distributions mixture of safety indicator in
  risk assessment problems,'' \emph{International Journal of System Assurance
  Engineering and Management}, pp. 1--9, 2019.

\bibitem{yimer2020study}
T.~H. Yimer, C.~Wen, X.~Yu, and C.~Jiang, ``A study of the minimum safe
  distance between human driven and driverless cars using safe distance
  model,'' 2020.

\end{thebibliography}

\begin{IEEEbiography}[{\includegraphics[width=1in,height=1.25in,clip,keepaspectratio]{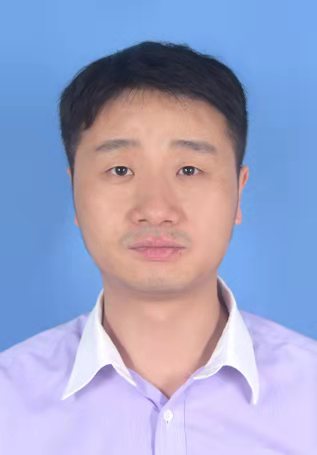}}]{Kafeng Wang}(S'21) is currently a Ph.D Student affiliated to Shenzhen Institute of Advanced Technology, Chinese Academy of Sciences and University of Chinese Academy of Sciences, Guangdong Province, China. He received his M.Sc Degree in Computer Sciences (2010) and B.Eng Degree in Software Engineering (2007), both from East China University of Technology, Nanchang, Jiangxi, China. His research interests include deep learning, artificial intelligence, and self-driving cars.
\end{IEEEbiography}

\begin{IEEEbiography}[{\includegraphics[width=1in,height=1.25in,clip,keepaspectratio]{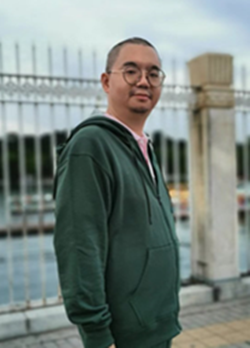}}]{Haoyi Xiong}
(S'11--M’15) received the Ph.D. degree in computer science from Telecom SudParis, Universit{\'e} Pierre et Marie Curie, Paris, France, in 2015. From 2016 to 2018, he was an Tenure-Track Assistant Professor with the Department of Computer Science, Missouri University of Science and Technology, Rolla, MO, USA (formerly known as University of Missouri at Rolla). From 2015 to 2016, he was a Post-Doctoral Research Associate with the Department of Systems and Information Engineering, University of Virginia, Charlottesville, VA, USA. He is currently a Principal R\&D Architect and Researcher with Big Data Laboratory, Baidu Research, Beijing, China.

His current research interests include automated deep learning (AutoDL), pervasive computing, and internet of things. He has published more than 60 papers in top computer science conferences and journals, such as ICML, ICLR, UbiComp, RTSS, AAAI, IJCAI, ICDM, PerCom, IEEE Transactions on Mobile Computing, IEEE Internet of Things Journal, IEEE Transactions on Neural Networks and Learning Systems, IEEE Transactions on Computers, ACM Transactions on Intelligent Systems and Technology, ACM Transactions on Knowledge Discovery from Databases, and etc. He gave keynote speeches in a series of academic and industrial activities, such as the industrial session of the 19$^{th}$ IEEE International Conference on Data Mining (ICDM'19), and served as Poster Co-chair for the 2019 IEEE International Conference on Big Data (IEEE Big Data'19). Dr. Xiong was a recipient of the Best Paper Award from IEEE UIC 2012, the Outstanding Ph.D. Thesis Runner Up Award from CNRS SAMOVAR 2015, and the Best Service Award from IEEE UIC 2017. He was one of the co-recipients of the prestigious Science \& Technology Advancement Award (First Prize) from Chinese Institute of Electronics 2019, and has been granted the prestigious IEEE TCSC Award for Excellence in Scalable Computing (Early Career Researcher), 2020 by IEEE Computer Society Technical Committee on Scalable Computing (IEEE TCSC).  
\end{IEEEbiography}

\begin{IEEEbiography}[{\includegraphics[width=1in,height=1.25in,clip,keepaspectratio]{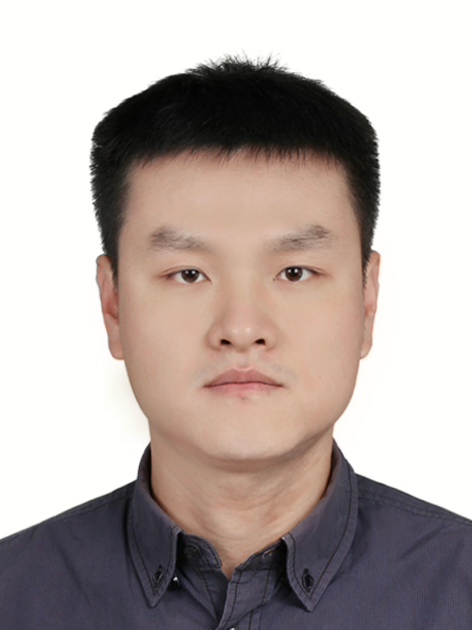}}]{Jie Zhang}(M'19) received the Ph.D. degree from University of Science and Technology Beijing, China, in 2019.
He was a visiting research fellow with University of Leeds, Leeds, U.K., in 2018. He is currently a postdoctoral research fellow with the School of Electronics Engineering and Computer Science, Peking University, Beijing, China. His research interests include machine learning, deep learning, and wireless sensor networks.
\end{IEEEbiography}

\begin{IEEEbiography}[{\includegraphics[width=1in,height=1.25in,clip,keepaspectratio]{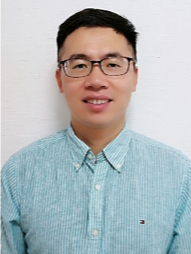}}]{Hongyang Chen}(SM'15) (Senior Member, IEEE) received the B.S. and M.S. degrees from Southwest Jiaotong University, Chengdu, China, in 2003 and 2006, respectively, and the Ph.D. degree from The University of Tokyo, Tokyo, Japan, in 2011. From 2004 to 2006, he was a Research Assistant with the Institute of Computing Technology, Chinese Academy of Science, Beijing, China. In 2009, he was a Visiting Researcher with Adaptive Systems Laboratory, University of California, Los Angeles, Los Angeles, CA, USA. From 2011 to 2020, he was a Researcher with Fujitsu Ltd., Tokyo, Japan. He is currently a Senior Research Expert with Zhejiang Lab, China. He has authored or coauthored 100 refereed journal and conference papers in the ACM Transactions on Sensor Networks, the IEEE TRANSACTIONS ON SIGNAL PROCESSING, the IEEE TRANSACTIONS ON WIRELESS COMMUNICATIONS, the IEEE MILCOM, the IEEE GlOBECOM, and the IEEE ICC, and has been granted or filed more than 50 PCT patents. His research interests include IoT, data-driven intelligent networking and systems, machine learning, localization, location-based big data, B5G, and statistical signal processing. He was a Symposium Chair or Special Session Organizer for some flagship conferences, including the IEEE PIMRC, IEEE MILCOM, IEEE GLOBECOM, and IEEE ICC. He was the recipient of the Best Paper Award from the IEEE PIMRC’09. He was the Editor of the IEEE TRANSACTIONS ON WIRELESS COMMUNICATIONS and the Associate Editor for the IEEE COMMUNICATIONS LETTERS. He is currently a leading Guest Editor of the IEEE JOURNAL ON SELECTED TOPICS OF SIGNAL PROCESSING on tensor decomposition and an Associate Editor for the IEEE INTERNET OF THINGS. He has been selected as the Distinguished Lecturer of the IEEE Communication Society from 2021 to 2022.
\end{IEEEbiography}

\begin{IEEEbiography}[{\includegraphics[width=1in,height=1.25in,clip,keepaspectratio]{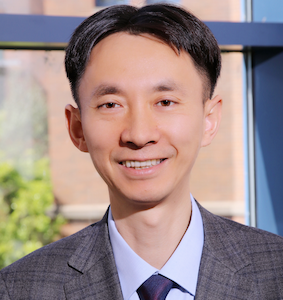}}]{Dejing Dou}(SM'20) is currently on the sabbatical leave from the Computer and Information Science Department at the University of Oregon, Eugene, OR, USA, where he also leads the Advanced Integration and Mining (AIM) Lab. He serves as the Director of the NSF IUCRC Center for Big Learning (CBL). He received his bachelor degree from Tsinghua University, China in 1996 and his Ph.D. degree from Yale University in 2004. His research areas include artificial intelligence, data mining, data integration, information extraction, and health informatics. Dejing Dou has published more than 100 research papers, some of which appear in prestigious conferences and journals like AAAI, IJCAI, KDD, ICDM, ACL, EMNLP, CIKM, ISWC, JIIS and JoDS. His DEXA'15 paper received the best paper award. His KDD'07 paper was nominated for the best research paper award. He is on the Editorial Boards of Journal on Data Semantics, Journal of Intelligent Information Systems, and PLOS ONE. He has been serving as program committee members for various international conferences and as program co-chairs for four of them. Dejing Dou has received over \$5 million PI research grants from the NSF and the NIH. He was promoted to Full Professor in 2016. He is currently the Head of Big Data Laboratory, Baidu Research, Beijing, China.
\end{IEEEbiography}

\begin{IEEEbiography}[{\includegraphics[width=1in,height=1.25in,clip,keepaspectratio]{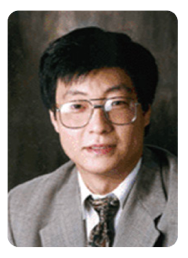}}]{Chengzhong Xu}
(F'16) received the PhD degree from the University of Hong Kong, in 1993. He is the Dean of Faculty of Science and Technology and the Interim Director of Institute of Collaborative Innovation, University of Macau, and a Chair Professor of Computer and Information Science. He was a professor of Wayne State University and the Director of Institute of Advanced Computing of Shenzhen Institutes of Advanced Technology, Chinese Academy of Sciences before he joined UM in 2019. Dr. Xu is a Chief Scientist of Key Project on Smart City of MOST, China and a Principal Investigator of the Key Project on Autonomous Driving of FDCT, Macau SAR. Dr. Xu's main research interests lie in parallel and distributed computing and cloud computing, in particular, with an emphasis on resource management for system's performance, reliability, availability, power efficiency, and security, and in big data and data-driven intelligence applications in smart city and self-driving vehicles.  He has published more than 200 papers in journals and conferences. He serves on a number of journal editorial boards,including the IEEE Transactions on Computers, the IEEE Transactions on Parallel and Distributed Systems, the IEEE Transactions on Cloud Computing, the Journal of Parallel and Distributed Computing and the China Science Information Sciences. He is a Fellow of the IEEE. \end{IEEEbiography}

\end{document}